\pdfoutput=1
\documentclass{article}
\usepackage{spconf,amsmath,amssymb,graphicx}
\usepackage{booktabs}
\usepackage{multirow}
\usepackage{hyperref}
\usepackage{url}
\usepackage{enumitem}
\usepackage[table]{xcolor}
\usepackage{placeins}
\usepackage{bbding}
\usepackage{algorithm2e}
\usepackage{makecell}
\usepackage{adjustbox}
\usepackage{tabularx}
\usepackage{cite}

\title{FPI-Det: A Face--Phone Interaction Dataset for Phone-Use Detection and Understanding}

\name{Jianqin Gao$^{1}$ \; Tianqi Wang$^{2}$ \; Yu Zhang$^{3}$ \; Yishu Zhang$^{2}$ \; Chenyuan Wang$^{2}$ \; Allan Dong$^{1}$ \; Zihao Wang$^{1}$%
\thanks{Jianqin Gao and Tianqi Wang contributed equally (co-first authors).\protect\\
Corresponding author: Zihao Wang (\texttt{Zihao.wang@sydney.edu.au}).}}%

\address{$^{1}$ The University of Sydney \\
         $^{2}$ Taiyuan University of Technology \\
         $^{3}$ Xi'an University of Science and Technology}

\begin{document}
\ninept
\maketitle

\begin{abstract}
The widespread use of mobile devices has created new challenges for vision systems in safety monitoring, workplace productivity assessment, and attention management. Detecting whether a person is using a phone requires not only object recognition but also an understanding of behavioral context, which involves reasoning about the relationship between faces, hands, and devices under diverse conditions. Existing generic benchmarks do not fully capture such fine-grained human--device interactions. To address this gap, we introduce the FPI-Det, containing 22{,}879 images with synchronized annotations for faces and phones across workplace, education, transportation, and public scenarios. The dataset features extreme scale variation, frequent occlusions, and varied capture conditions. We evaluate representative YOLO and DETR detectors, providing baseline results and an analysis of performance across object sizes, occlusion levels, and environments. Source code and dataset is available at \url{https://github.com/KvCgRv/FPI-Det}.
\end{abstract}

\section{Introduction}

The increasing ubiquity of mobile devices has introduced significant challenges for applications requiring behavior-aware visual perception. In particular, determining whether a person is using a mobile phone is an essential task in domains such as traffic safety, workplace monitoring, education, and human–computer interaction. This is not merely an object detection problem but a critical component of situational awareness systems. For instance, in transportation, driver distraction due to mobile phone use is a leading cause of traffic accidents, making automated detection a life-saving technology. In educational settings, identifying off-task phone usage can help maintain an engaged learning environment. Unlike conventional object detection problems, this task requires not only recognizing the presence of phones and faces but also understanding their subtle and dynamic contextual relationships, including facial orientation, gaze direction, hand positioning, and other interaction cues within complex, often cluttered environments.

Prior work typically frames mobile-phone use detection as a binary image-classification task and employs Convolutional Neural Network (CNN) backbones such as Residual Network-50 (ResNet-50) \cite{he2016deep}. While convenient for coarse screening, this formulation collapses diverse behaviors into two labels and cannot capture fine-grained cues—e.g., face orientation, gaze, and hand–device configurations. In contrast, modern object detectors can jointly localize faces and phones and reason about their spatial relationships, yielding more precise behavioral understanding and enabling lightweight deployment on edge devices. A number of studies have endeavored to develop algorithms for mobile phone usage detection in different situations ~\cite{fu2024gd,zhao2024yolo,poon2021driver,zhao2024behavior,liu2024yolo,wang2024mf}; however, the available datasets are predominantly manually curated and annotated, and they lack unified standards or benchmarks.

Traditional general-purpose object detection benchmarks such as Common Objects in Context (COCO) \cite{coco} and the PASCAL Visual Object Classes (PASCAL VOC) \cite{pascal} have driven remarkable advances in generic detection, but they primarily emphasize isolated object categories and do not capture fine-grained human–device interactions. Meanwhile, face detection datasets such as WIDER FACE \cite{widerface} and the Face Detection Data Set and Benchmark (FDDB) \cite{fddb} focus on robustness to variations in pose, scale, and occlusion, yet they lack synchronized annotations of contextual devices such as mobile phones. Other datasets for human pose estimation or interaction recognition, including the Max Planck Institute for Informatics (MPII) Human Pose \cite{mpii} and Humans Interacting with Common Objects (HICO-DET) \cite{hico}, provide broader perspectives on human activity but do not address the specific and highly relevant scenario of phone usage. As a result, existing resources fall short when modeling both human attributes and interactive devices in a unified detection framework, creating a significant research gap: the lack of a specialized, large-scale dataset designed to train and evaluate models on the nuanced task of human–phone interaction.

\begin{figure}[ht]
  \centering
  \includegraphics[width=\linewidth]{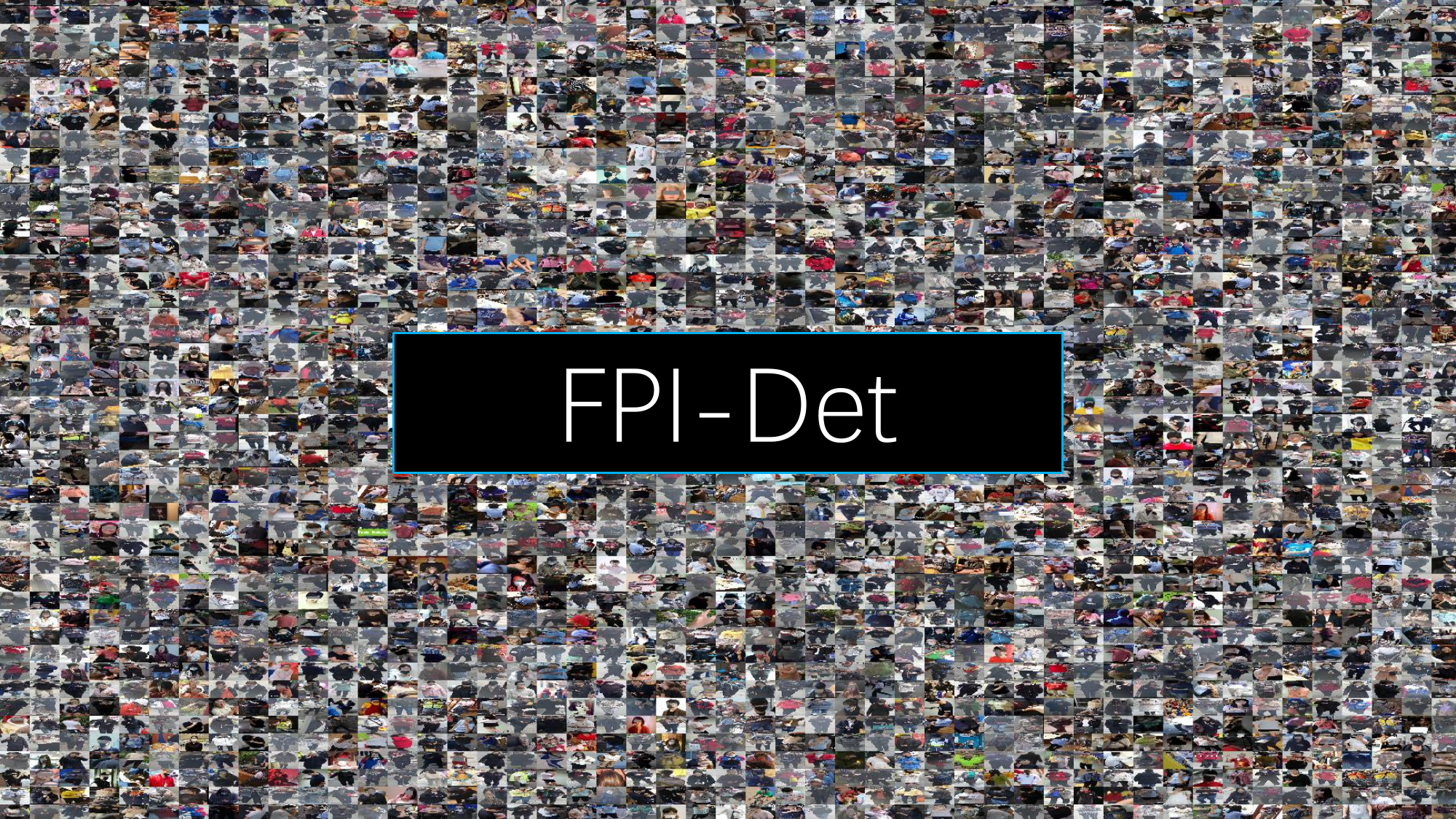}
  \caption{Overall scenario of FPI-Det.}
  \label{fig:main}
\end{figure}

To address this limitation and bridge this gap, we introduce FPI-Det, as shown in Figure~\ref{fig:main}, which provides synchronized annotations of faces and phones across diverse scenarios, including workplace, education, transportation, and public spaces. The dataset highlights three inherent and intertwined challenges. First, there is substantial scale variation, with phones often appearing as small as $20 \times 20$ pixels while faces may exceed $800 \times 800$ pixels. Second, mutual occlusion is frequent, where hands partially cover phones or obscure facial regions, increasing the difficulty of detection. Third, capture conditions are highly diverse, ranging from low-light environments and compression artifacts from surveillance footage to motion blur from handheld recording. Furthermore, the dataset captures the ambiguity of interaction; models must learn to differentiate between passive holding and active usage, a fine-grained distinction that pushes the boundaries of conventional detection. These factors make FPI-Det a challenging yet realistic benchmark for evaluating robust detection systems under real-world conditions.

In addition to the dataset itself, this paper makes several key contributions. We introduce FPI-Det as the first large-scale benchmark with synchronized, fine-grained annotations for faces, phones, and their interaction status. Concurrently, we establish a comprehensive evaluation protocol tailored for this task, providing standardized metrics to assess model performance. Finally, we conduct a comparative evaluation of representative detection paradigms. This analysis contrasts leading one-stage detectors such as the You Only Look Once (YOLO) family \cite{yolov4, yolov7, yolov8}, dominant in real-time applications, with modern transformer-based approaches like the DEtection TRansformer (DETR) \cite{carion2020end} and Deformable DETR \cite{zhu2020deformable}, which emphasize relational reasoning and global context. These models have rarely been systematically assessed in the context of fine-grained human–device interactions. 

% Through this study, we aim to establish FPI-Det as a foundational benchmark for advancing research in behavior-aware detection and to provide new insights into the strengths and weaknesses of different architectural paradigms when applied to this challenging yet practically significant task.

\section{Proposed Dataset}
\subsection{Dataset Scenarios}
The FPI-Det consists of 22,879 annotated images that capture people in daily activities across four primary domains: workplace, education, transportation, and public spaces. These domains were selected to reflect realistic contexts where mobile phone usage frequently occurs and may lead to safety-critical consequences. For instance, in workplace and educational environments, individuals may use phones while engaged in other tasks, whereas in transportation and public spaces, distractions can have direct implications for personal and public safety.

\begin{figure}[t]
  \centering
  % 小宏：一列里上下各一张，底部标注名称
  \newcommand{\paircol}[3]{%
    \begin{minipage}[b]{0.24\linewidth}
      \centering
      \includegraphics[width=\linewidth]{#1}\\[2pt]
      \includegraphics[width=\linewidth]{#2}\\[4pt]
      \small #3
    \end{minipage}%
  }
  \paircol{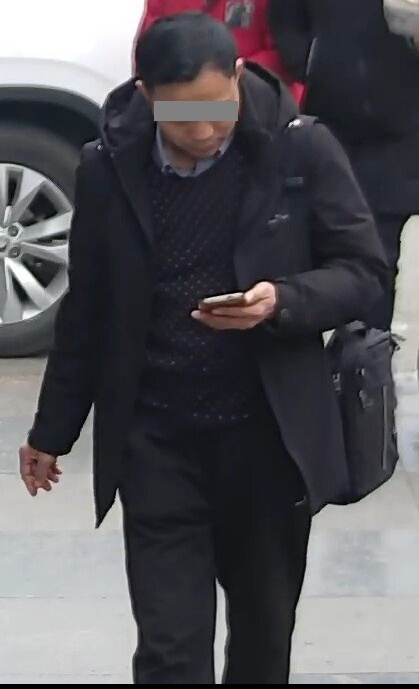}{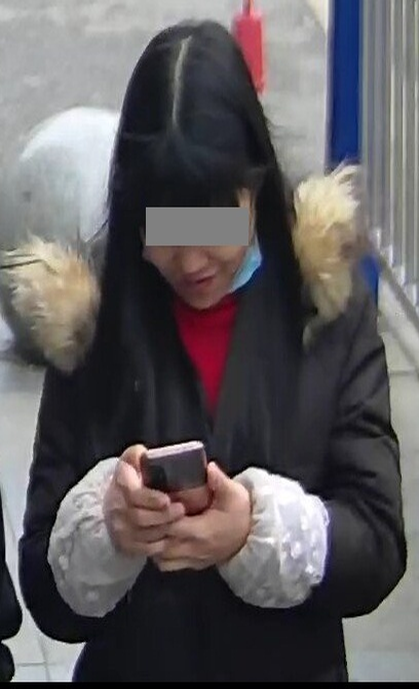}{\textit{using}}\hfill
  \paircol{"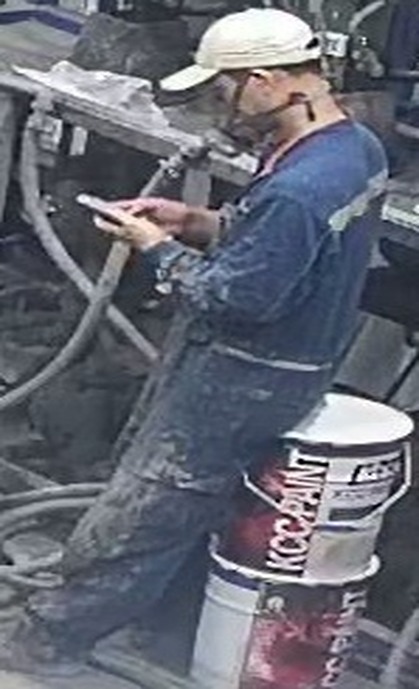"}{"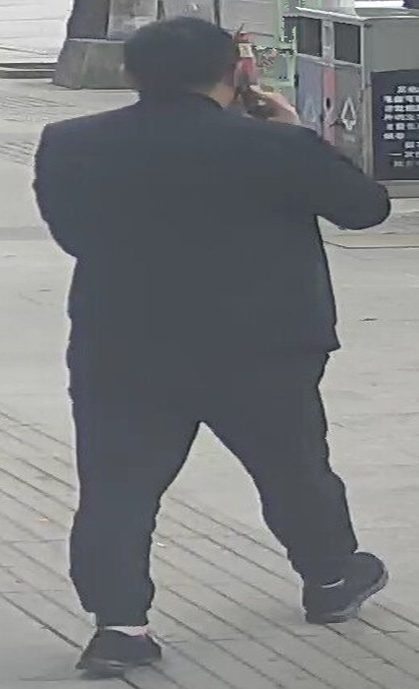"}{\textit{sidewise using}}\hfill
  \paircol{"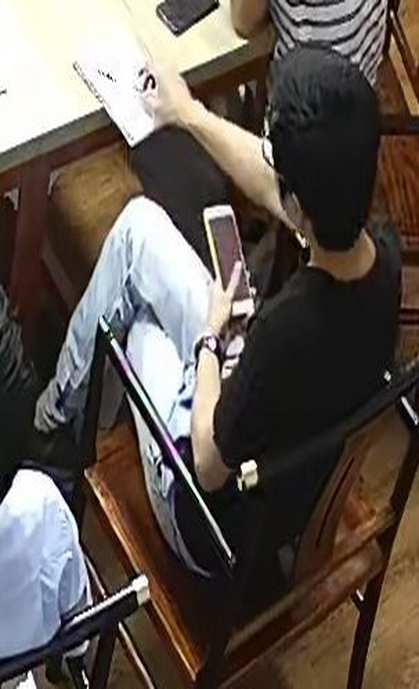"}{"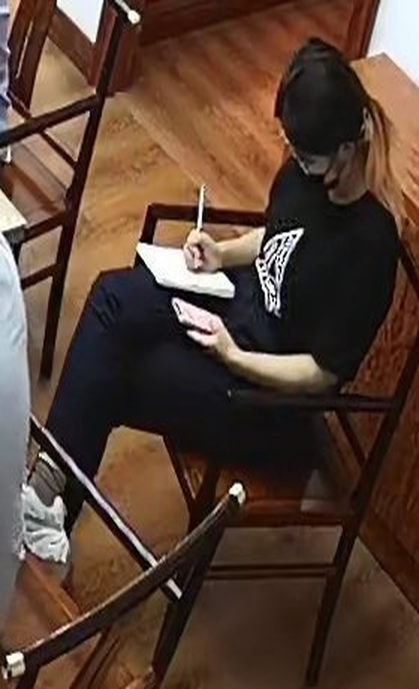"}{\textit{seated using  }}\hfill
  \paircol{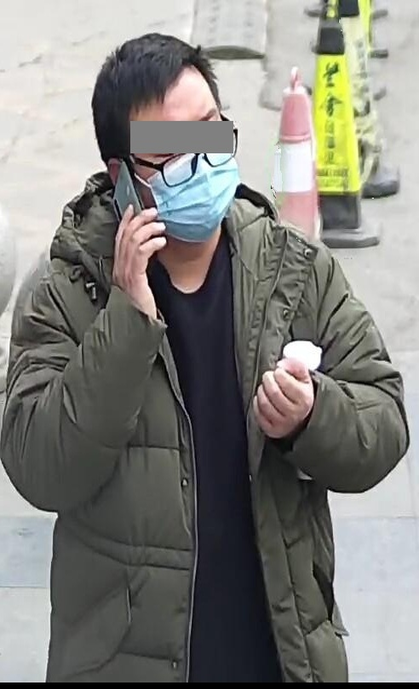}{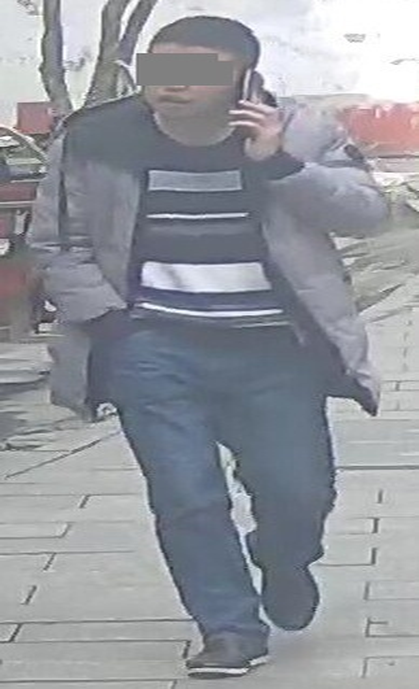}{\textit{calling}}
\caption{Paired exemplars for four actions: \textit{playing}, \textit{side playing}, \textit{sit playing}, and \textit{calling}.}

  \label{fig:fourcols_pairs}
\end{figure}

Figure~\ref{fig:fourcols_pairs} illustrates fine-grained categories of mobile phone usage behaviors in our dataset. Except for merely holding a phone without any interaction, the four behaviors shown are all treated as active smartphone use by the human subject. Example images from FPI-Det across different domains are also presented.

% 需要 \usepackage{graphicx}

\subsection{Dataset Construction}
The FPI-Det was constructed through a multi-stage process to ensure diversity and annotation quality. Images were collected from multiple sources, including surveillance cameras, mobile devices, and online video platforms, covering a wide variety of environments and recording conditions. Each image was manually inspected, and objects of interest—faces and mobile phones—were annotated with bounding boxes.

Given the complexity of phone–face interactions, the annotation process required careful handling of occlusion cases. Annotators were instructed to label visible phones even if they were partially covered by hands, and to include faces despite challenging conditions such as side views or partial visibility. To ensure annotation reliability, each bounding box was reviewed by an additional annotator, and discrepancies were resolved by consensus among at least three annotators.

Table~\ref{tab1} reports the counts and categories of all annotations in FPI-Det, whereas Table~\ref{tab2} presents the per-image breakdown. “Phone-Only” and “Face-Only” denote images that contain only a phone or only a human face, respectively. “Double” denotes images that contain both categories simultaneously, which is the most prevalent case in our dataset. “Null” denotes images without any annotations. The FPI-Det dataset includes 22,879 images,  with 29,279 faces and 10,255 phones annotated. Images contain on average 1.28 faces and 0.45 phones, with as many as 37 faces in a single frame. Among these, 9,888 images contain both phones and faces (Double), 12,164 are face-only, 367 are phone-only, and 460 are null.
 The dataset is divided into training, validation, and testing subsets, consisting of 18,800, 1,730, and 2,349 images, respectively. This balanced split supports both model development and fair evaluation.

Additionally, we provide a CSV-formatted annotation file for evaluating the model’s final classification performance on the test and validation sets. The file was annotated by two experts and offers fine-grained labels indicating whether the smartphone-use behavior of individuals depicted in the images is genuine.
\begin{table}[htb]
\caption{Quantity of images and annotations of FPI-Det dataset.}
\label{tab1}
\centering
\begin{tabular}{lccc}
\toprule
\textbf{Type} & \textbf{Face} & \textbf{Phone} &\textbf{Images}\\
\midrule
Train  & 24,993 & 8,358  &18800\\
Val   & 1,318  & 795 &1,730\\
Test  & 2,968 & 1,102  &2,349\\
\midrule
\textbf{Total} &29,279&10,255 & 22,879 \\
\bottomrule
\end{tabular}
\end{table}

\begin{table}[htbp]

\centering
\caption{Each image-level annotation is accompanied by an analysis.}
\label{tab2}
\resizebox{!}{0.06\textwidth}{
\begin{tabular}{cccccc}
\toprule
\textbf{FPI-Det} & \textbf{Phone-Only} & \textbf{Face-Only} & \textbf{Double} & \textbf{Null}& \textbf{Total} \\
\midrule

Train & 11 & 10394 & 8347 &48 & 18800 \\

Val & 352 & 548 & 443 &387 & 1730 \\

Test & 4 & 1222 & 1098 &25 & 2349 \\
\midrule
Total&367&12164&9888&460&22879 \\
\bottomrule
\end{tabular}
}
\end{table}

\subsection{What's New for this Benchmark}
\begin{figure}[ht]
  \centering
  \includegraphics[width=\linewidth]{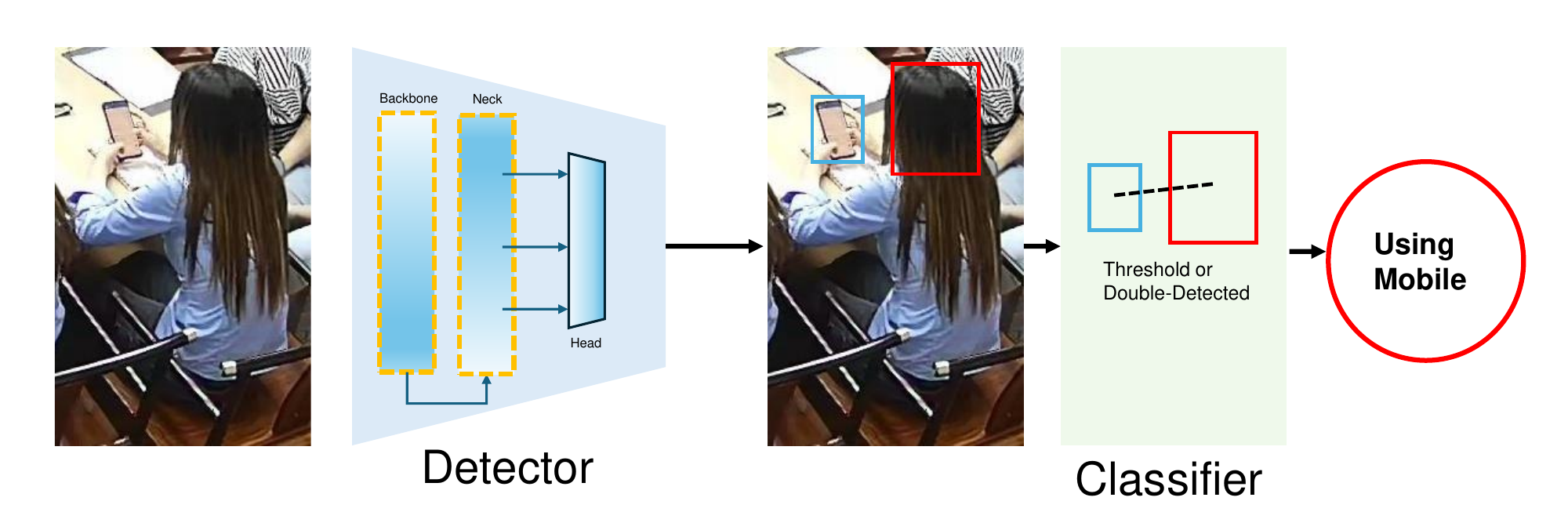}
  \caption{Pipeline of the two tasks of FPI-Det.}
  \label{fig:toy}
\end{figure}

Compared with the original data, we substantially expand face annotations and release both detection labels and a benchmark for binary phone-use classification with baseline results. To support fine-grained behavior inference, we provide per-instance geometry—pixel coordinates for faces and phones and their absolute pairwise distances—enabling distance-threshold rules and metric-learning formulations. We also annotate coarse behavior categories: Calling, Using,Seated using , and using obliquely. These additions enable new tasks: (i) phone-use detection in workplace settings, (ii) precise attribution of the phone-using subject in multi-person scenes, and (iii) a large-scale extension for face detection under surveillance-like viewpoints. The overall pipeline of our approached dataset is shown in Figure~\ref{fig:toy}.

\section{Experiments and Results}

To provide a thorough and fair evaluation of detection models on the FPI-Det, we designed and executed a rigorous set of experiments. This section details the selected architectures, evaluation protocols, and implementation environment, followed by a comprehensive analysis of the results.

\subsection{Experimental Setup}
We chose a representative set of models spanning two dominant paradigms: the efficient, one-stage YOLO series (YOLOv8 and YOLOv11) and the end-to-end transformer-based DETR and Deformable DETR. Model performance was assessed using mean Average Precision at IoU thresholds of 0.5 (mAP@50) and 0.95 (mAP@.95). For the YOLO family of models, we trained for 200 epochs with a batch size of 16. Given the convergence challenges of DETR-style detectors, we trained the DETR family for 300 epochs, using the same batch size of 16. We also analyzed results by object category, size, and occlusion level. To gauge practical viability, we measured inference speed in frames per second (FPS) on a single NVIDIA Tesla V100 GPU. All models were trained using identical hyperparameter configurations and data augmentation strategies to ensure fair comparison. Table~\ref{tab:env_config} details the hardware and software specifications of our experimental setup, provided to facilitate a rough estimation of the GPU training time required by researchers using the dataset. 
\begin{table}[!h]
\centering
\caption{Specifications of the experimental environment.}
\label{tab:env_config}
\begin{tabularx}{\columnwidth}{@{} p{0.4\columnwidth} X @{}}
\toprule
\textbf{Component} & \textbf{Specification} \\ 
\midrule
\multicolumn{2}{@{}l}{\textbf{Hardware}} \\
\quad GPU & 4 $\times$ NVIDIA Tesla V100  \\
\quad CPU & Intel Xeon Processor \\
\quad System Memory & 256 GB RAM \\ 
\midrule
\multicolumn{2}{@{}l}{\textbf{Software}} \\
\quad Operating System & Ubuntu 20.04 LTS \\
\quad CUDA Toolkit & 12.0 \\
\quad Framework & PyTorch 2.1 \\
\bottomrule
\end{tabularx}
\end{table}

\begin{figure*}[t]
\centering

\newcommand{\cellw}{0.11\textwidth}

\newcommand{\colthree}[4]{%
  \begin{minipage}[b]{\cellw}
    \centering
    \includegraphics[width=\linewidth]{#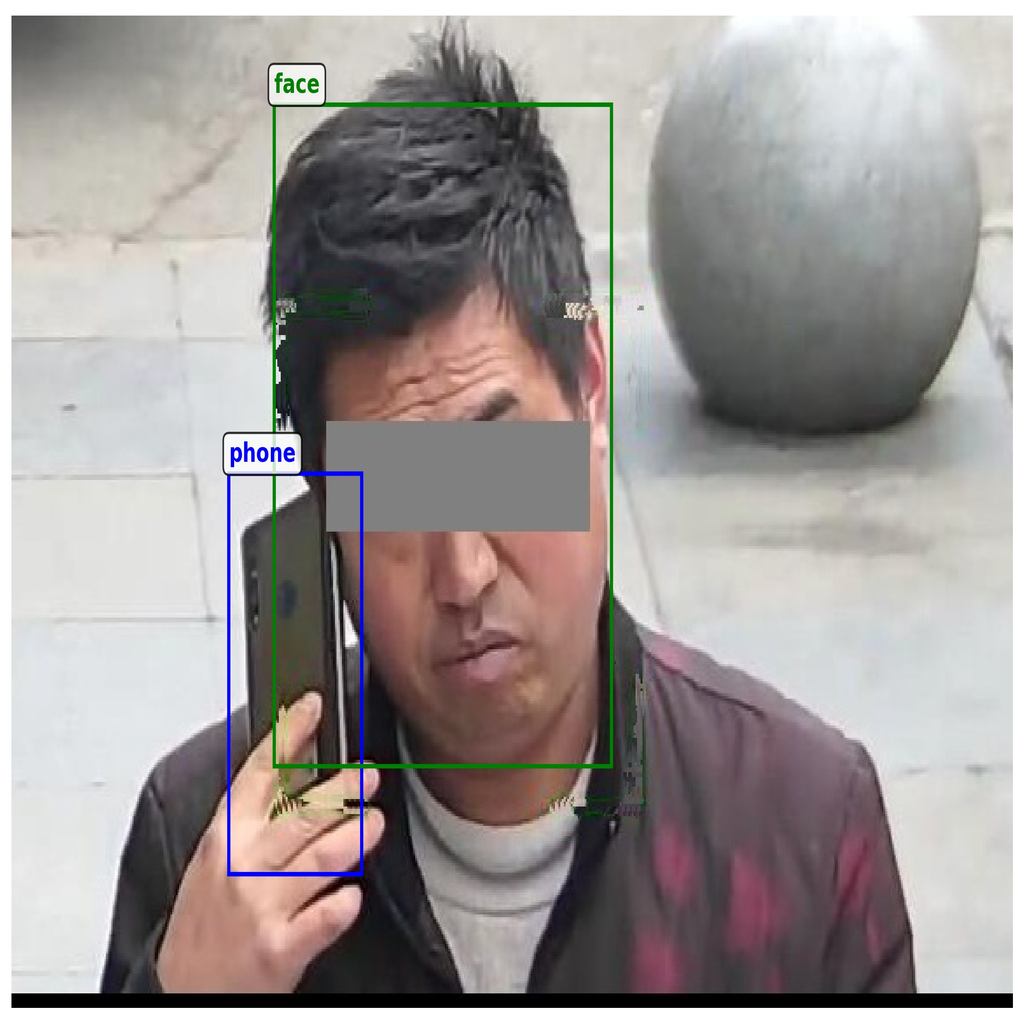}\\[2pt]
    \includegraphics[width=\linewidth]{#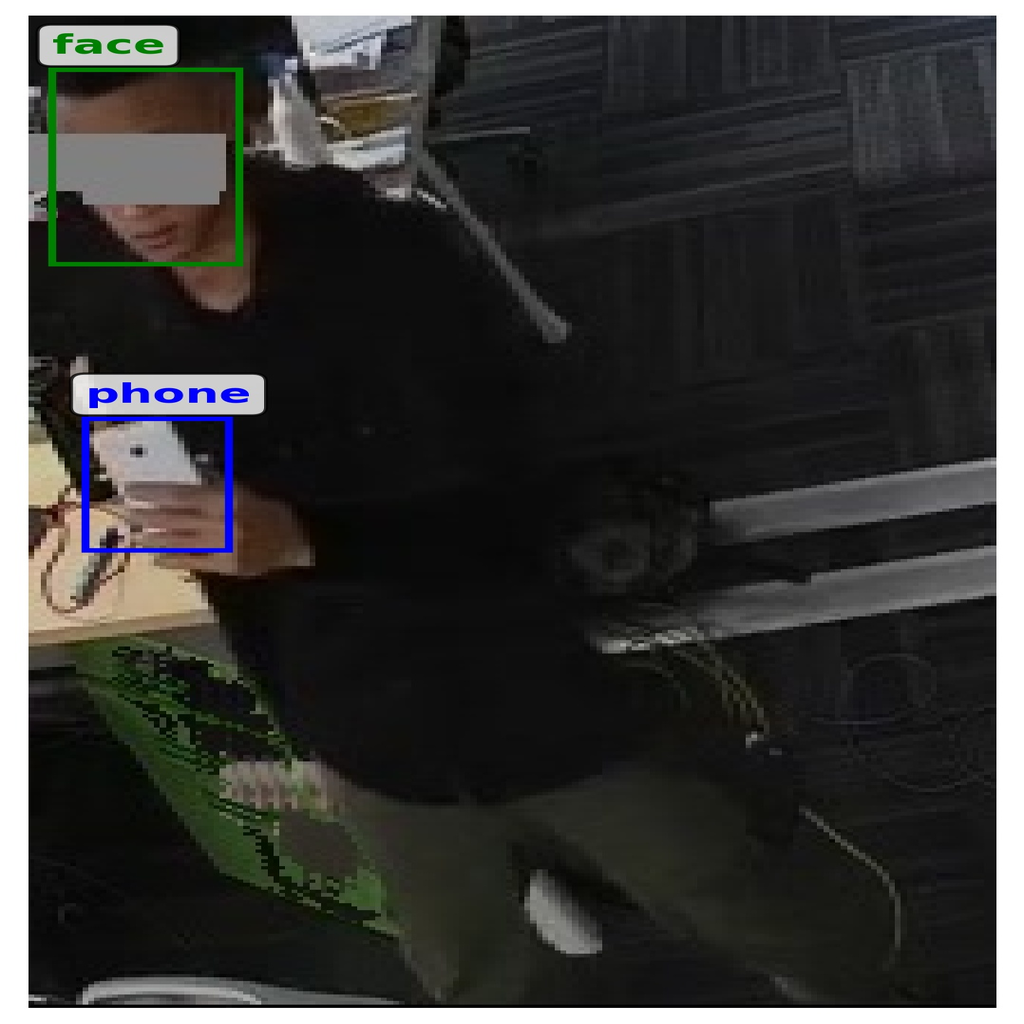}\\[2pt]
    \includegraphics[width=\linewidth]{#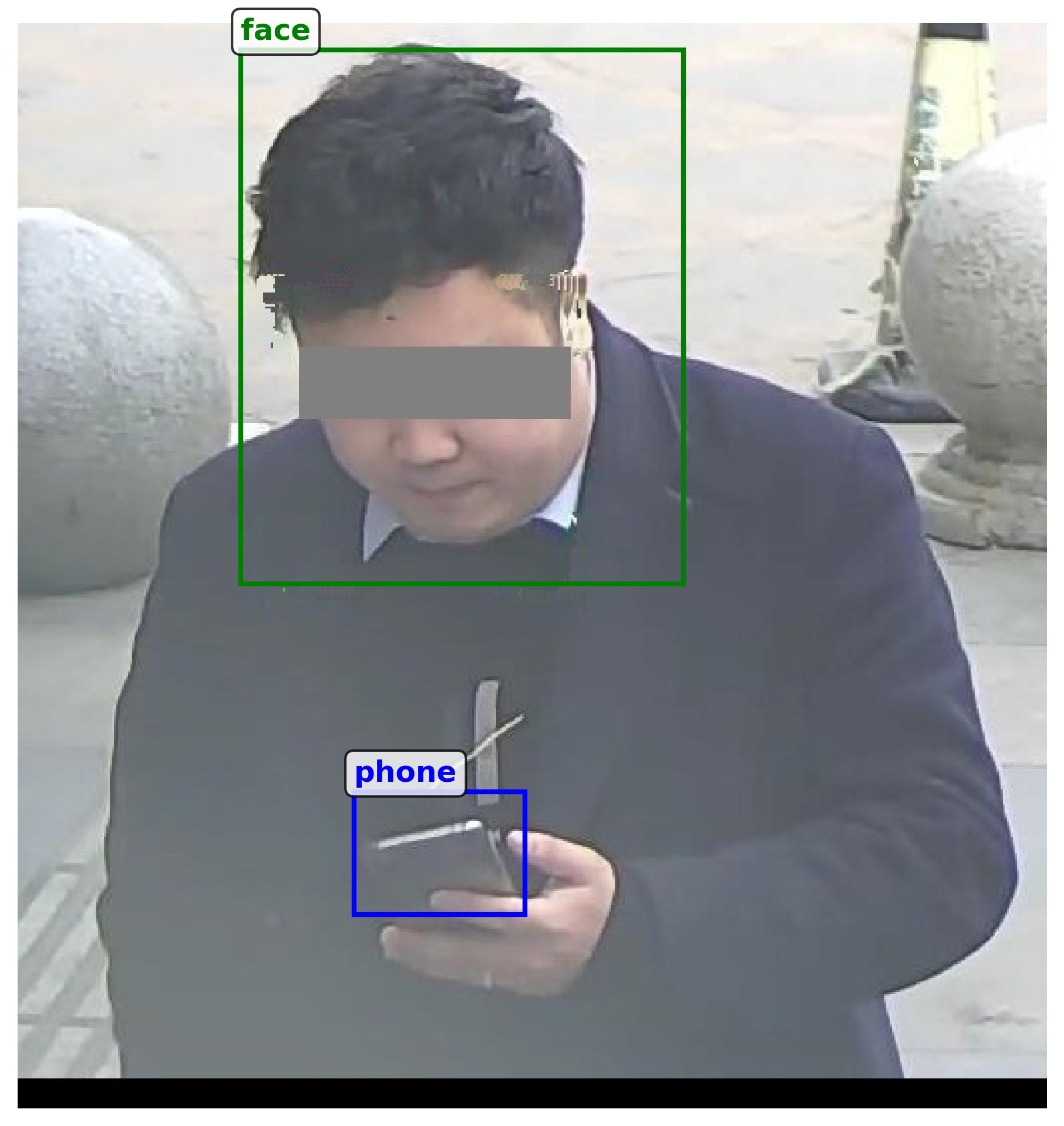}\\[-1pt]
    \scriptsize #4
  \end{minipage}%
}

\colthree{1}{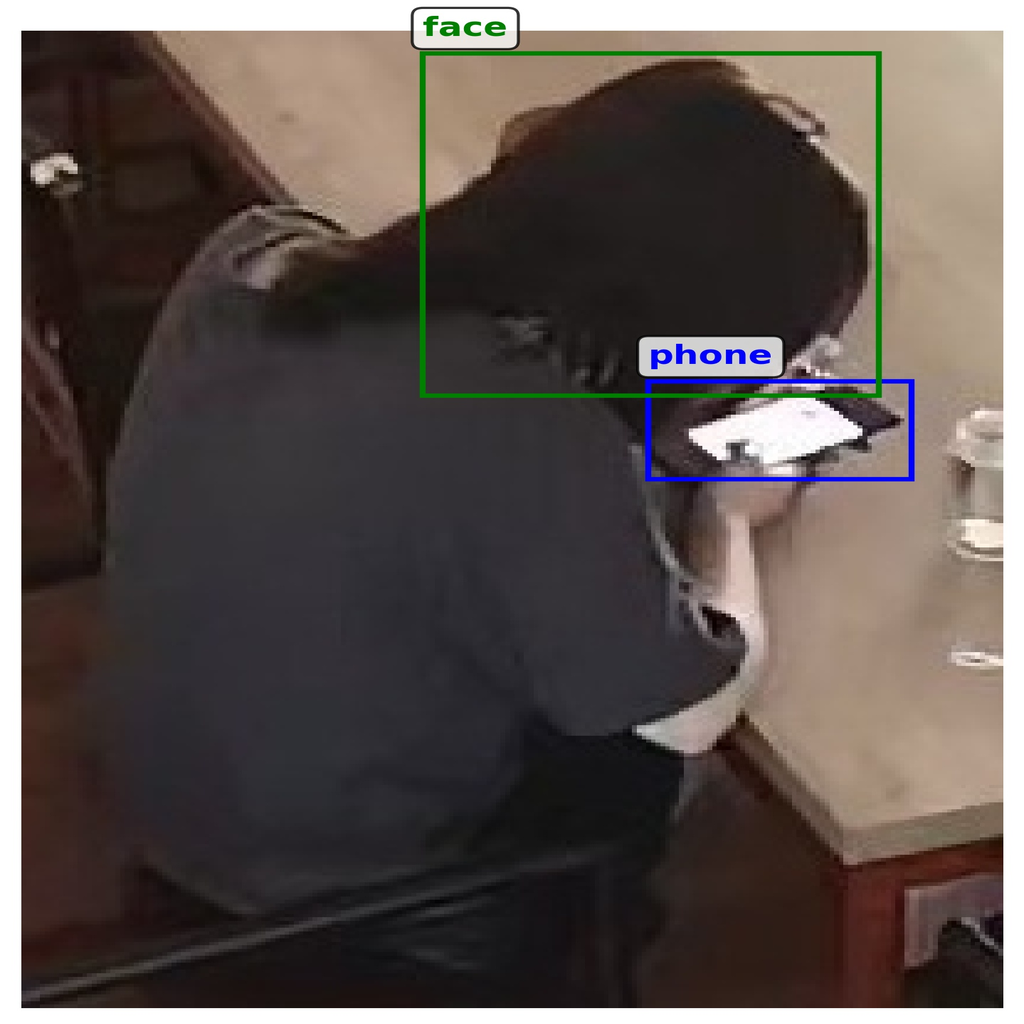}{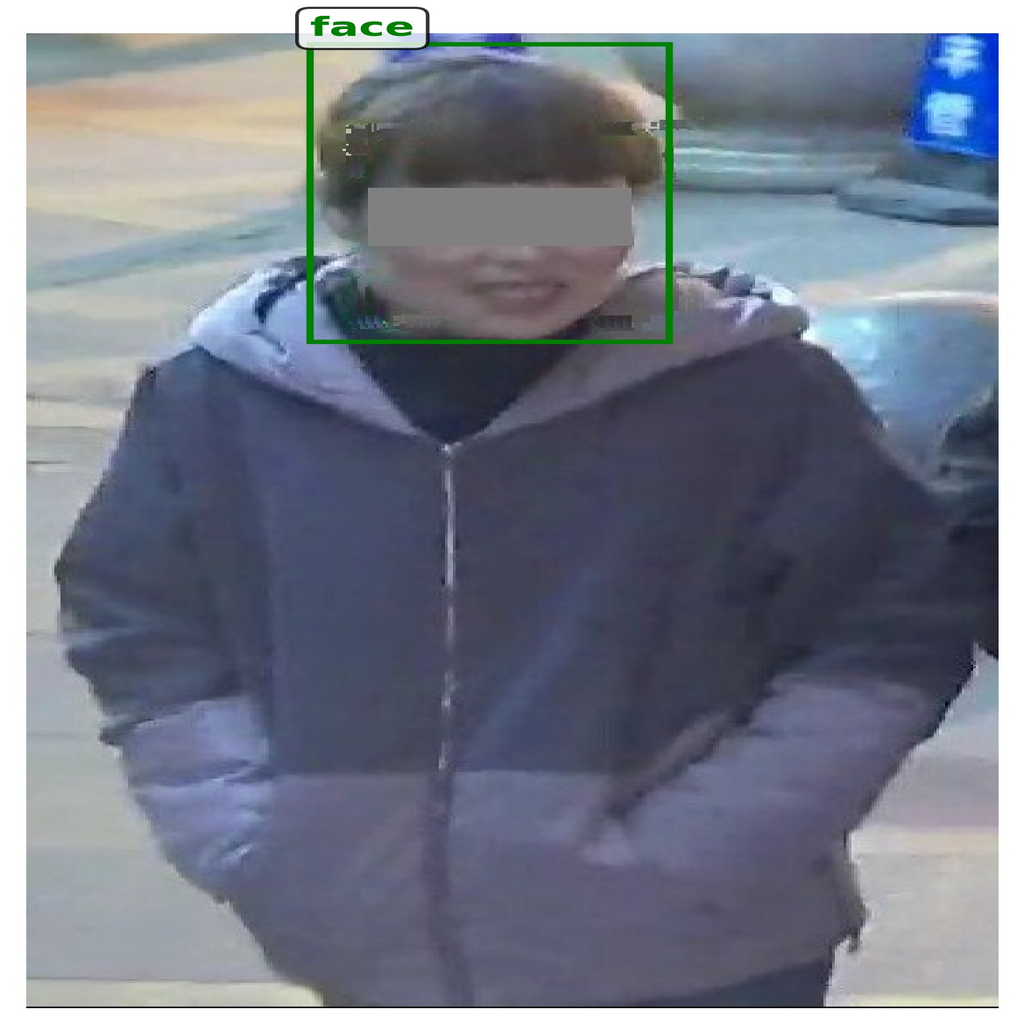}{YOLOv8-n}\hfill
\colthree{2}{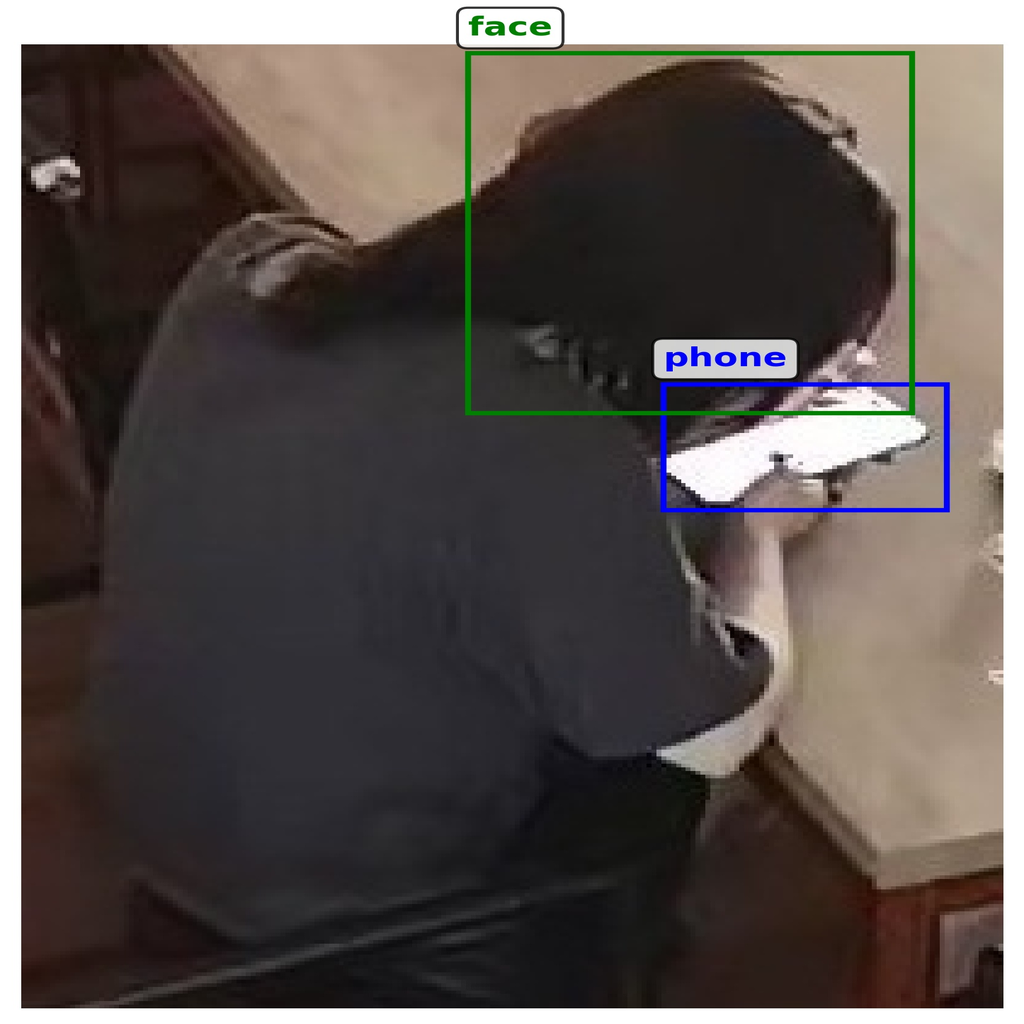}{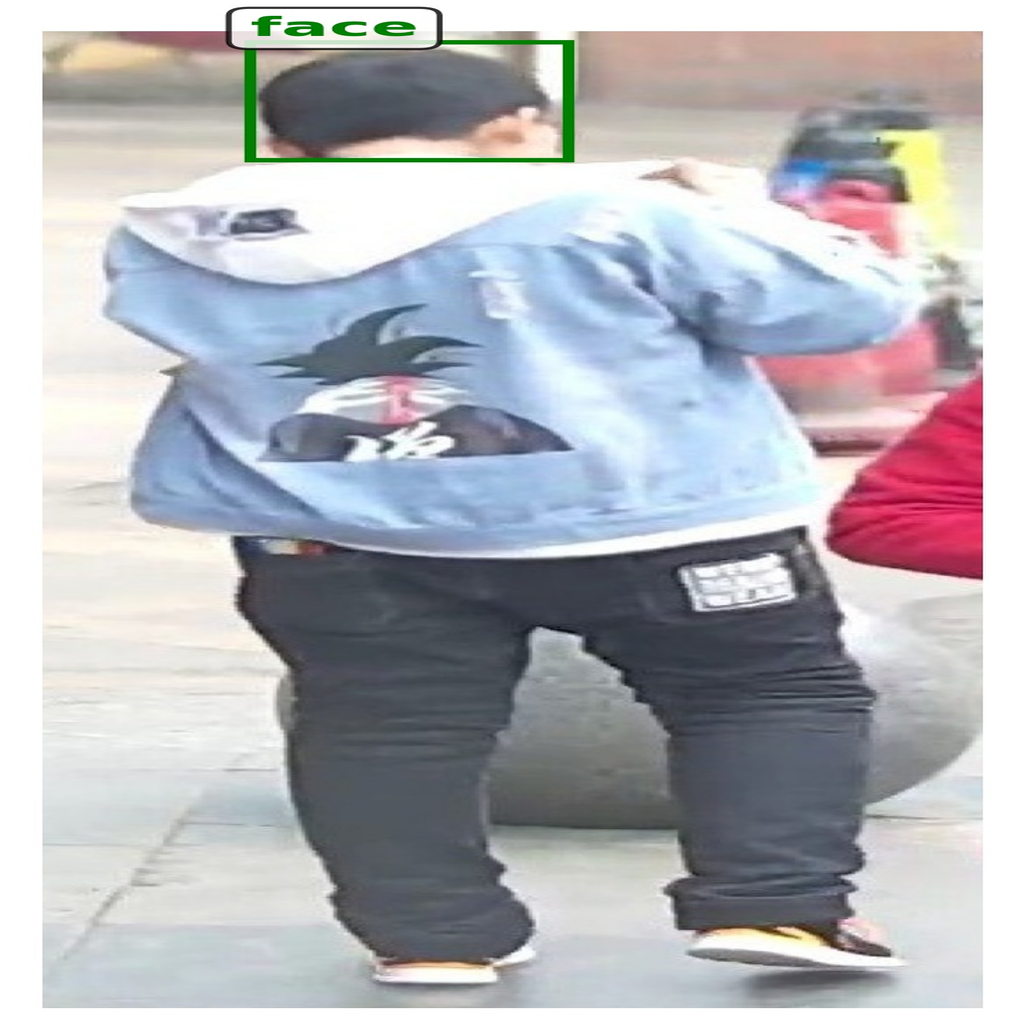}{YOLOv8-s}\hfill
\colthree{3}{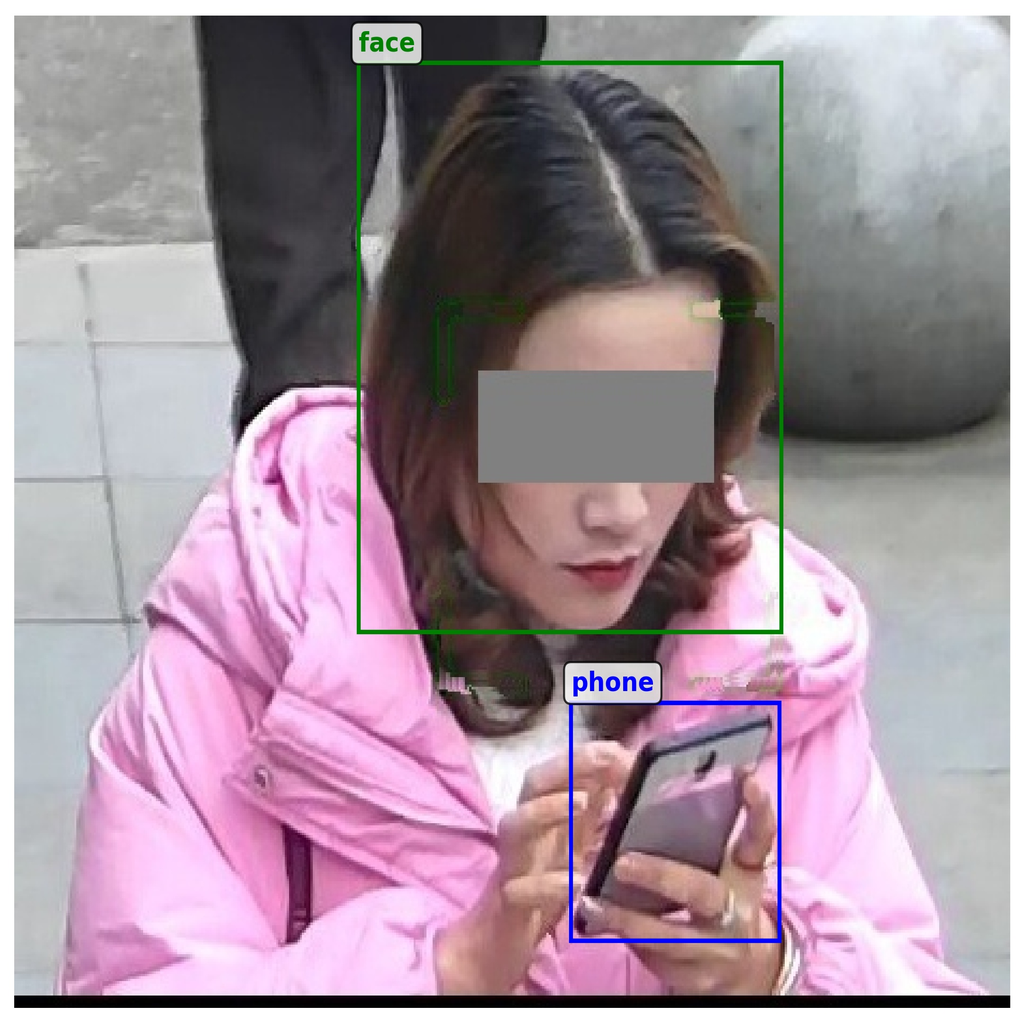}{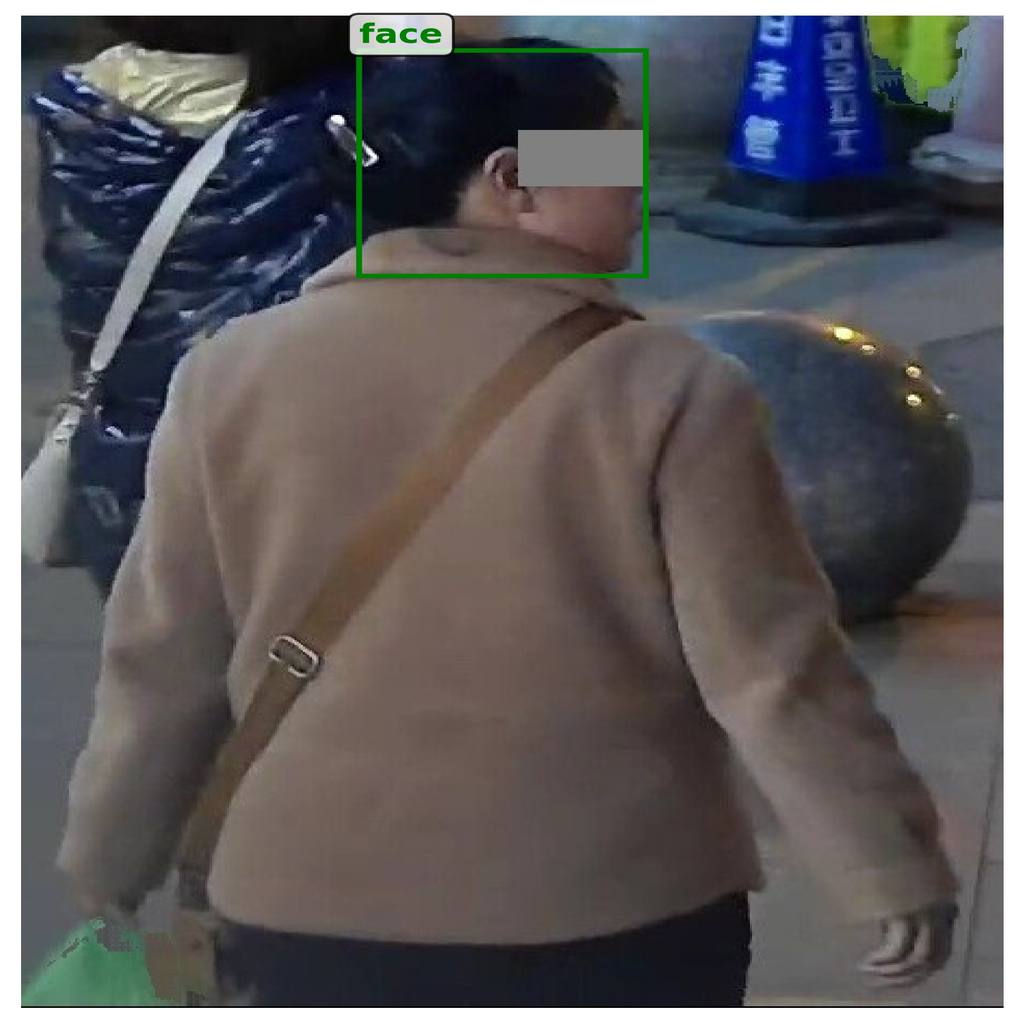}{YOLOv8-x}\hfill
\colthree{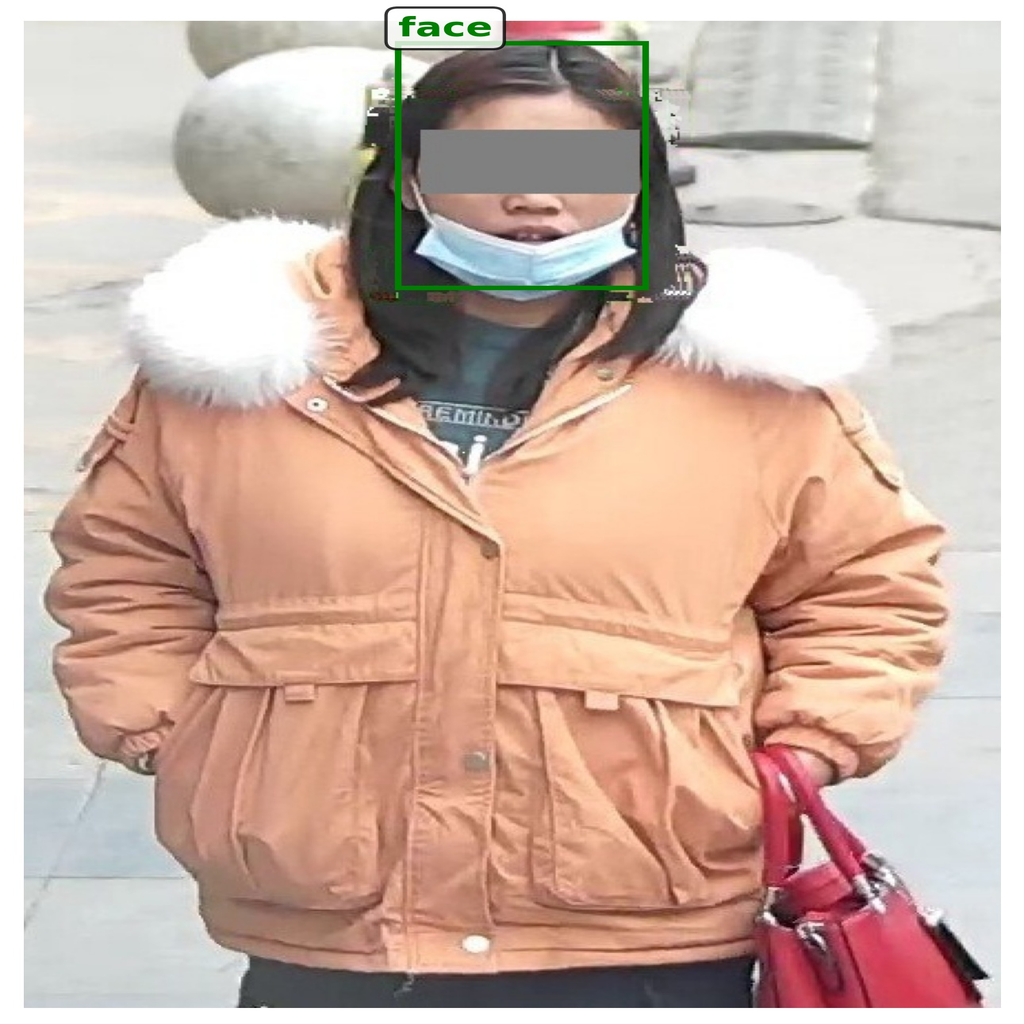}{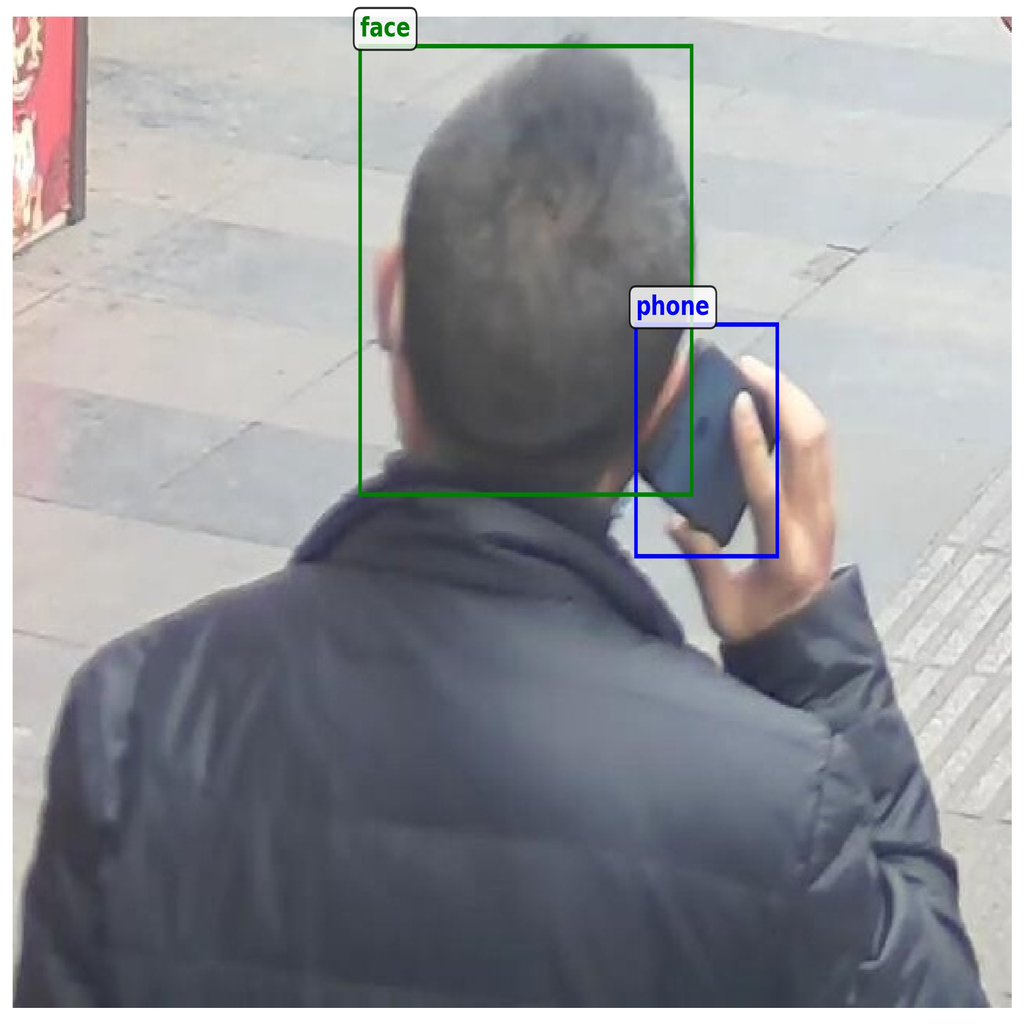}{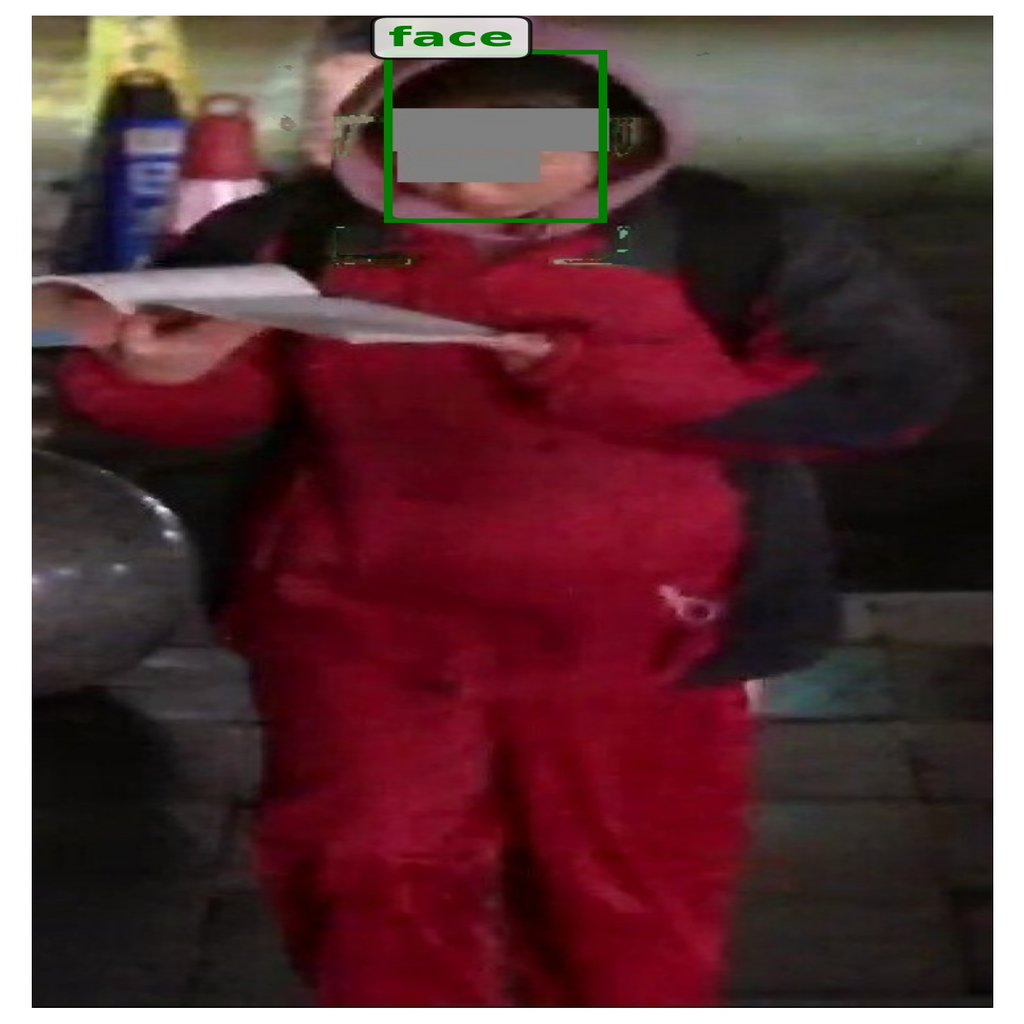}{YOLOv11-n}\hfill
\colthree{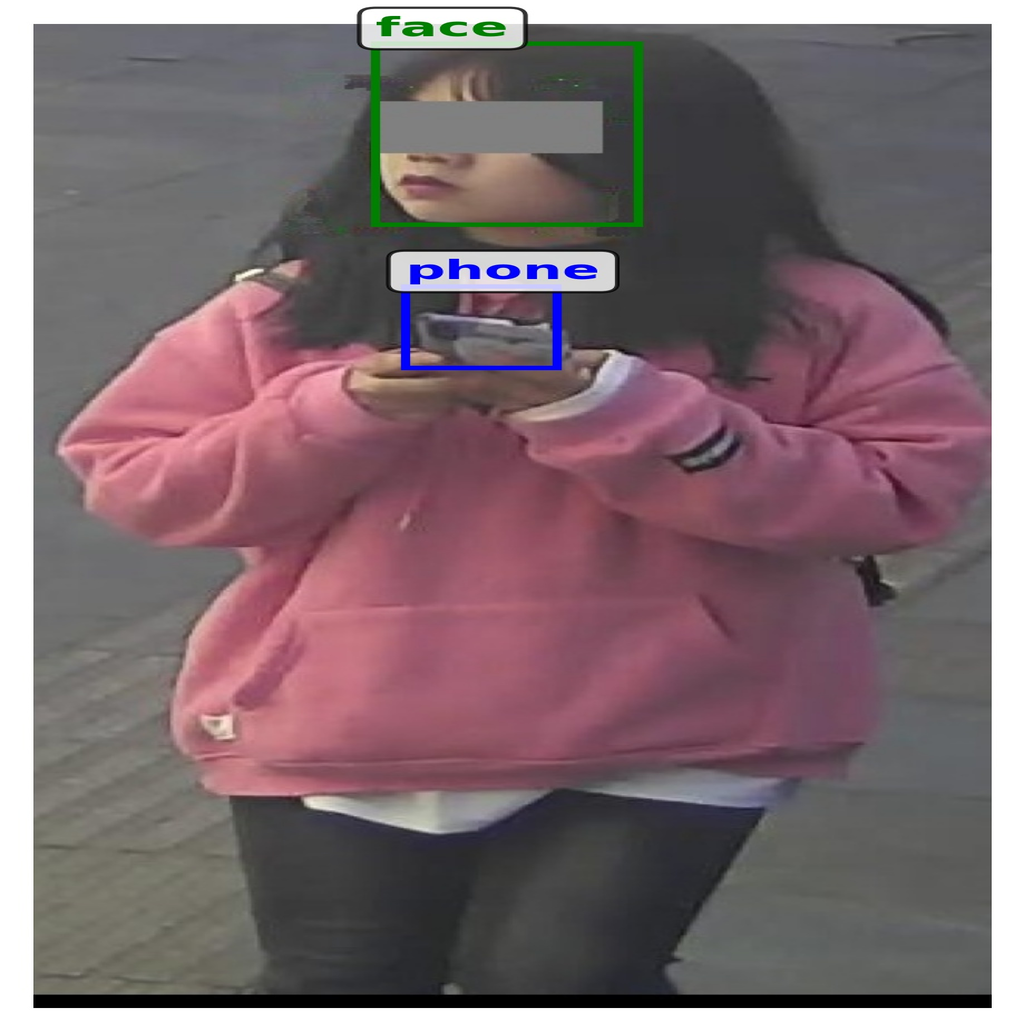}{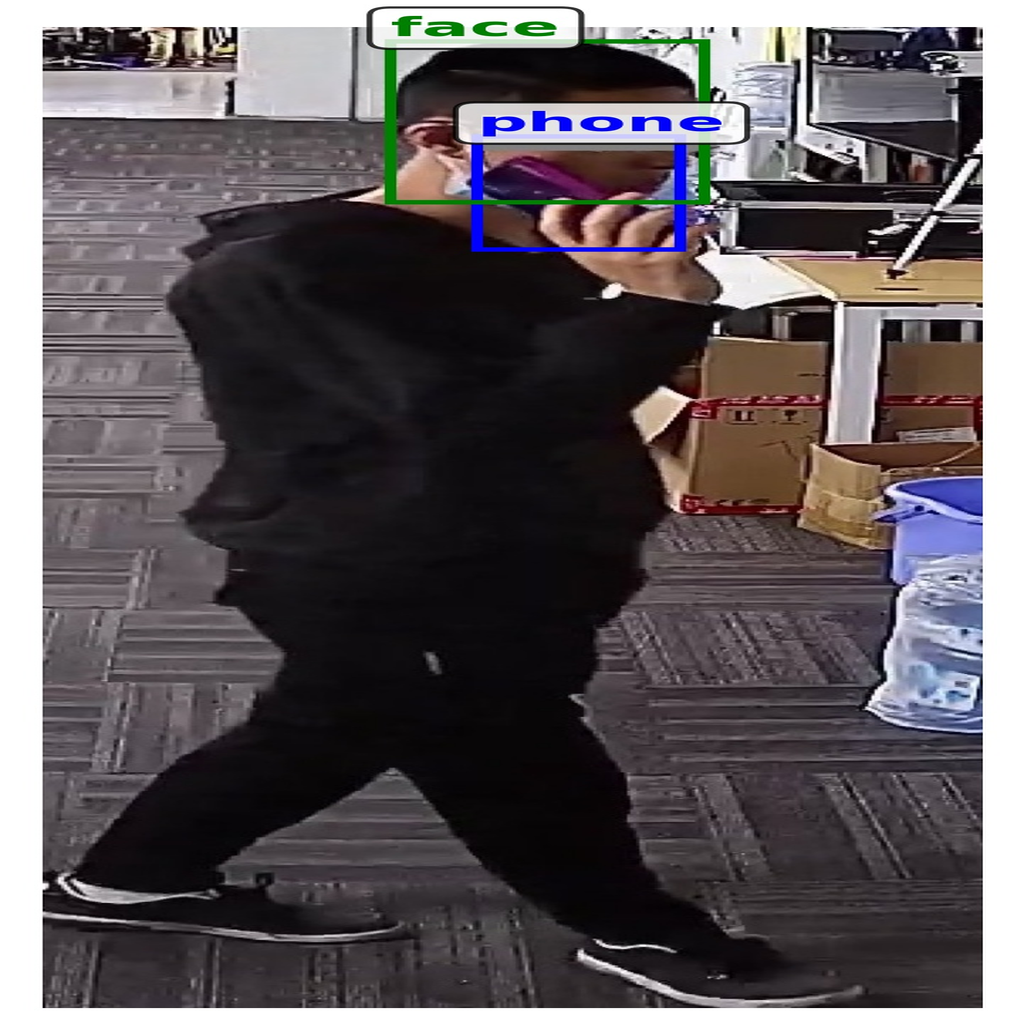}{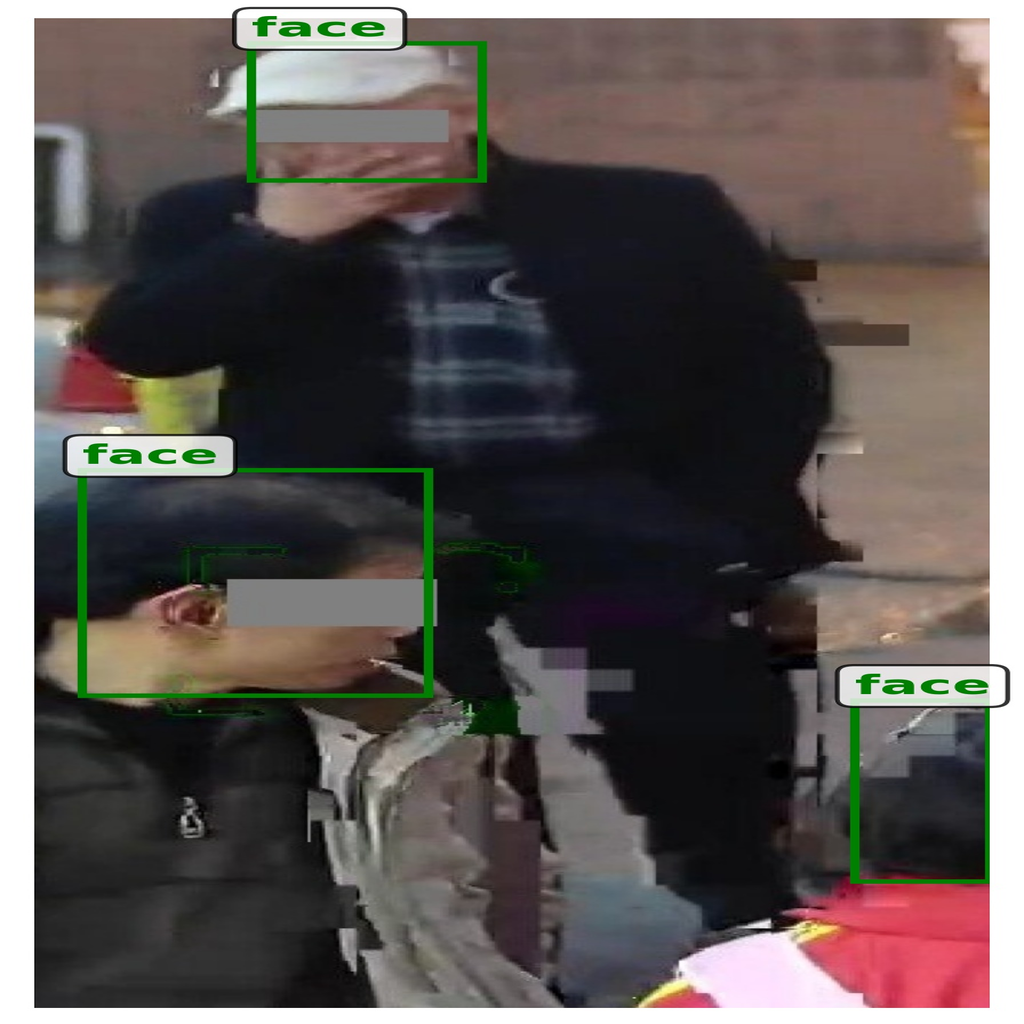}{YOLOv11-s}\hfill
\colthree{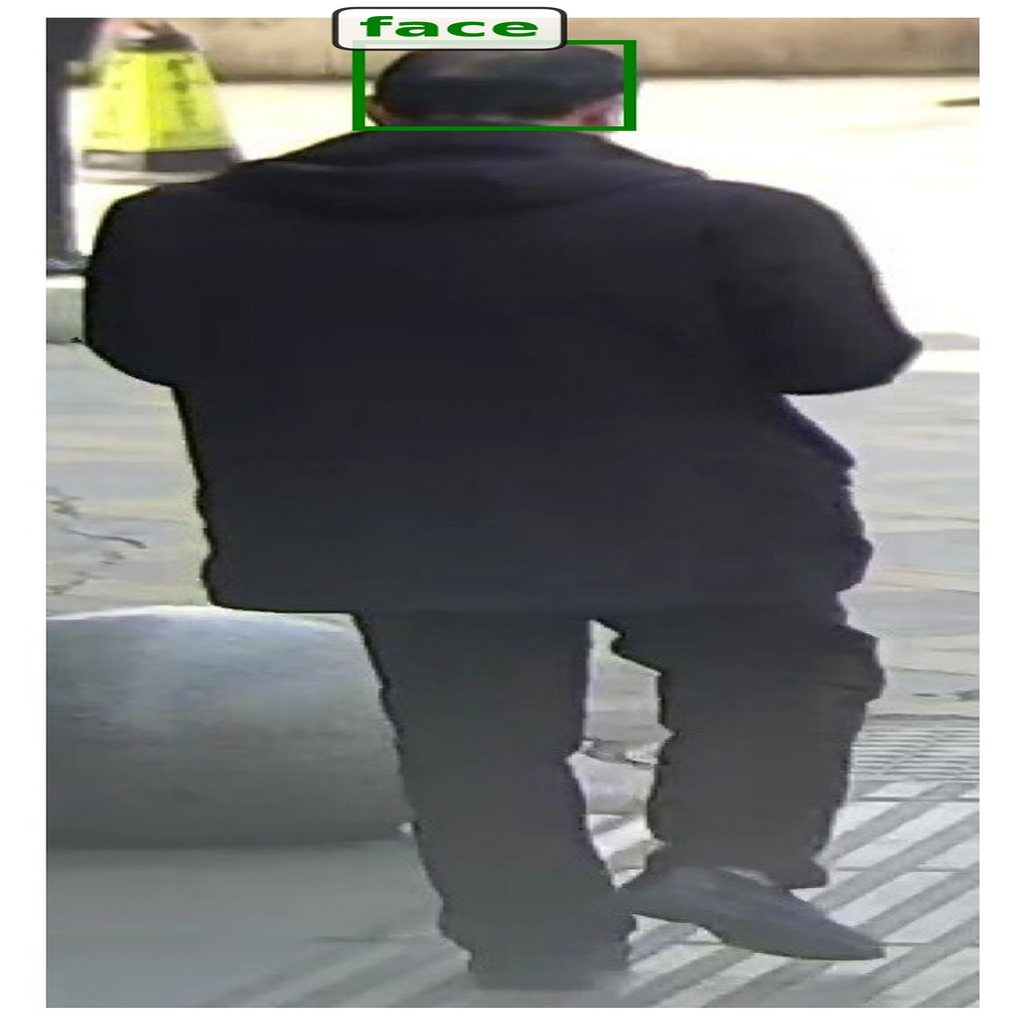}{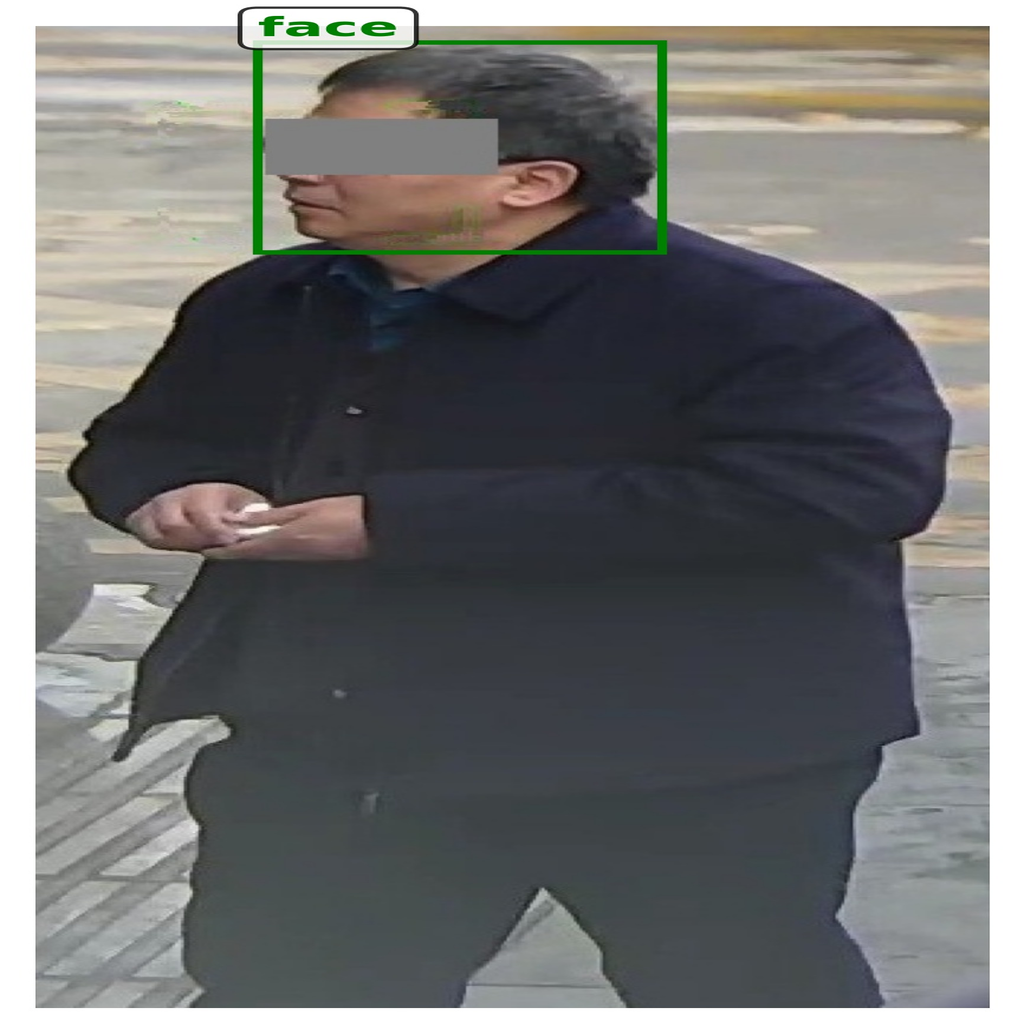}{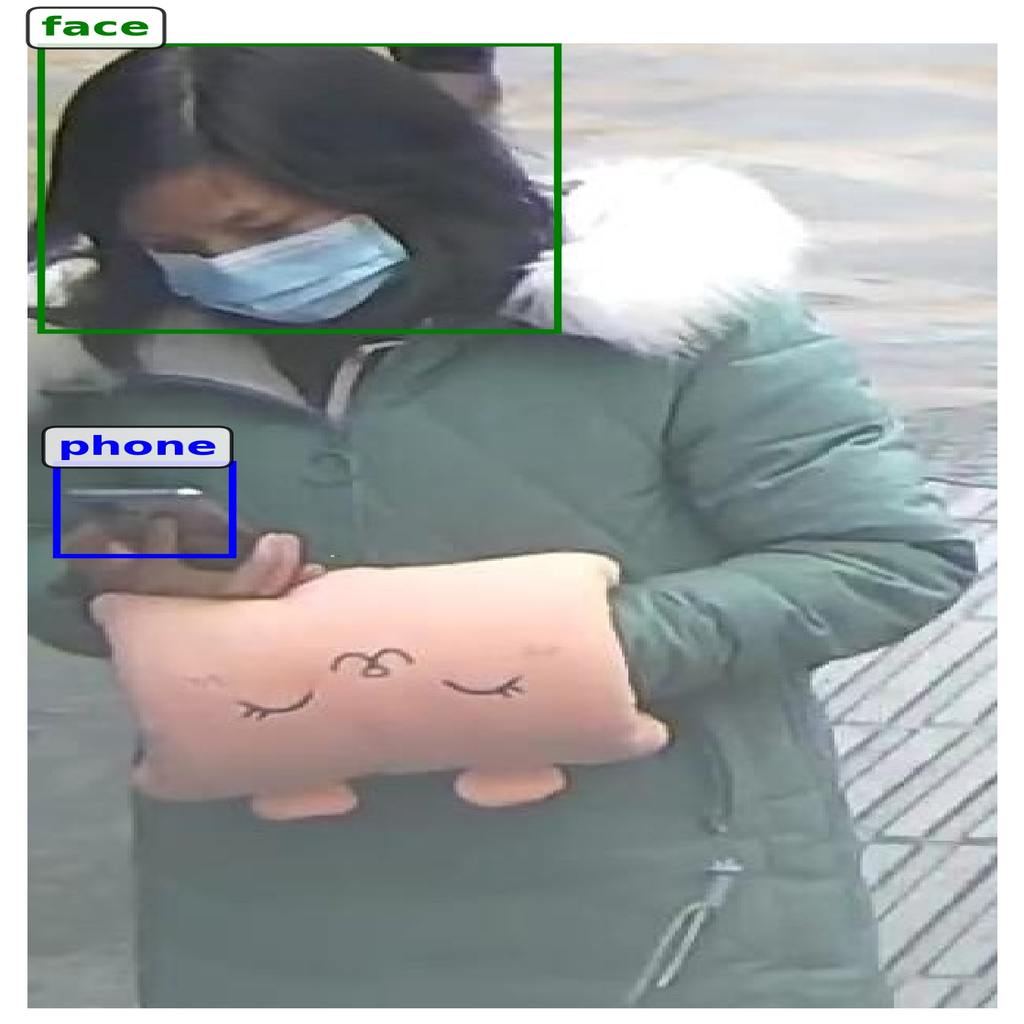}{YOLOv11-m}\hfill
\colthree{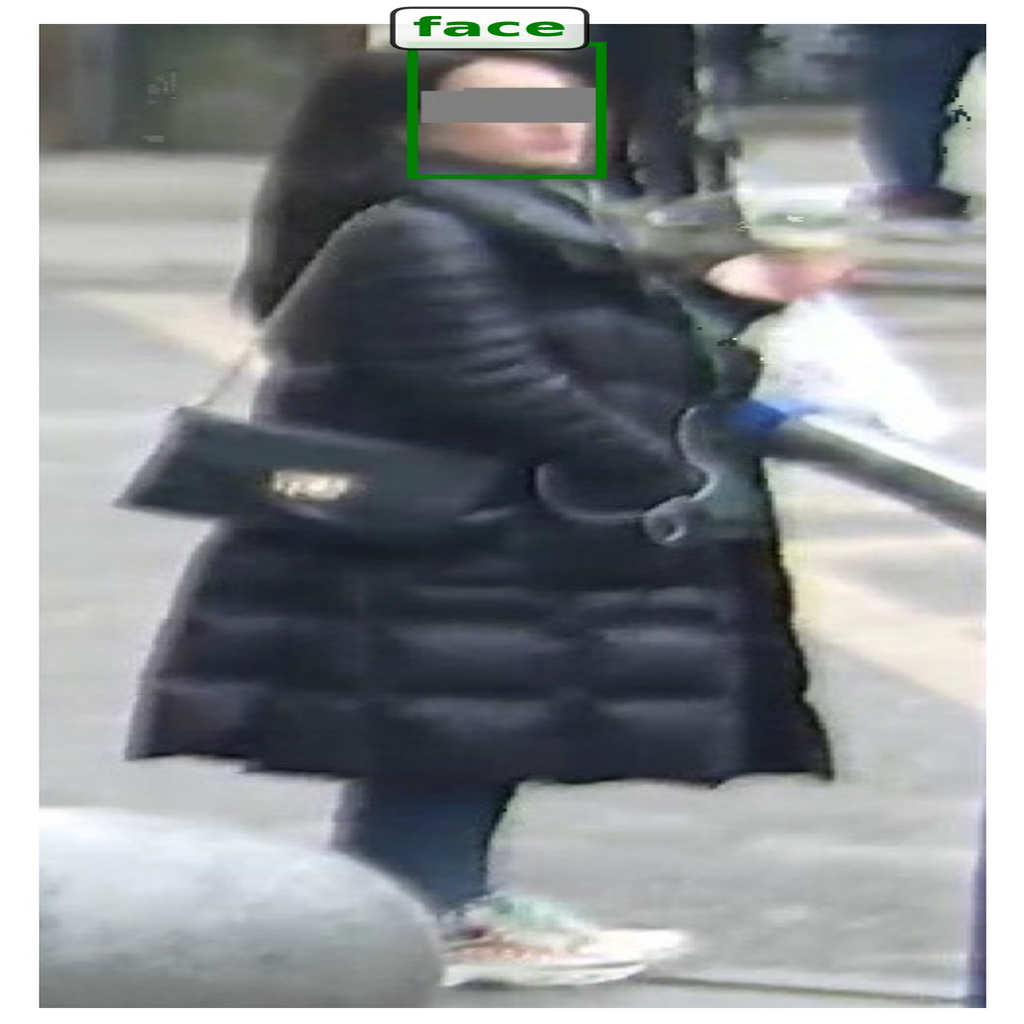}{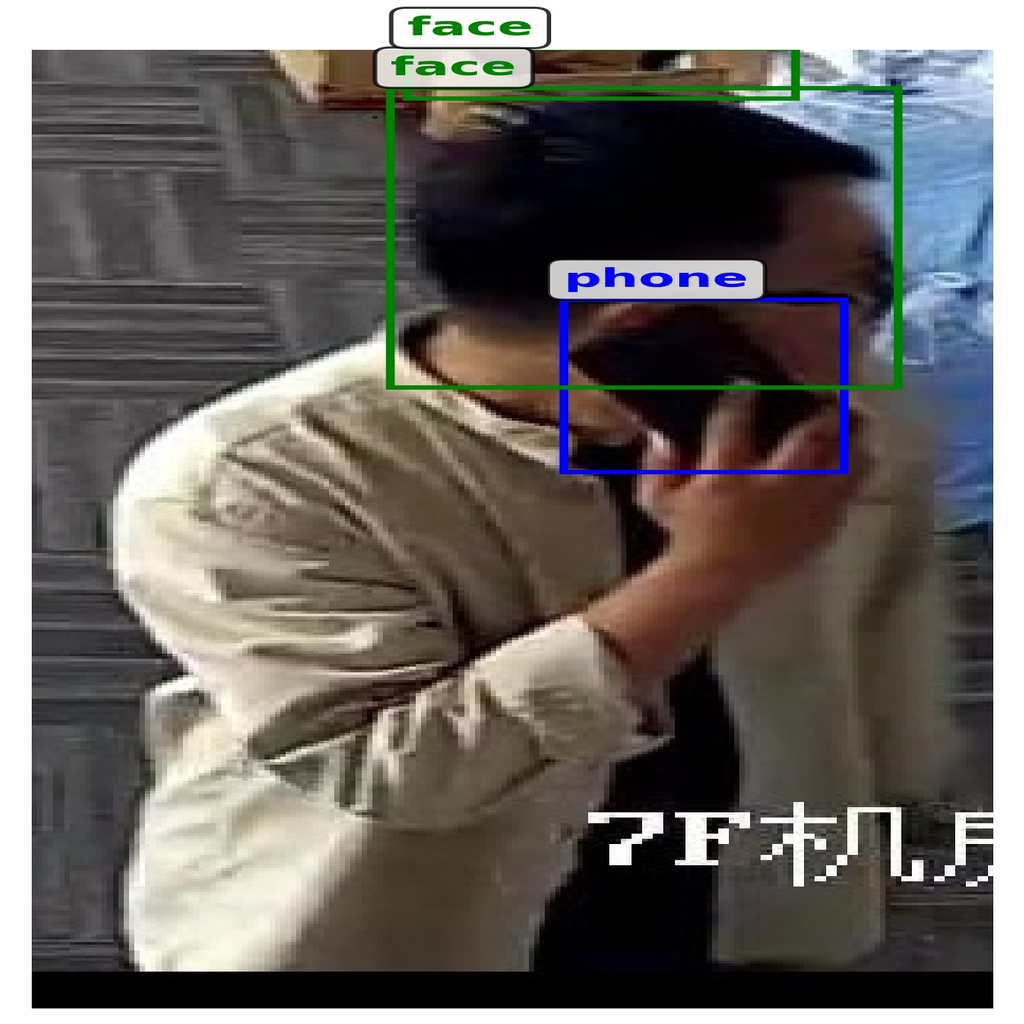}{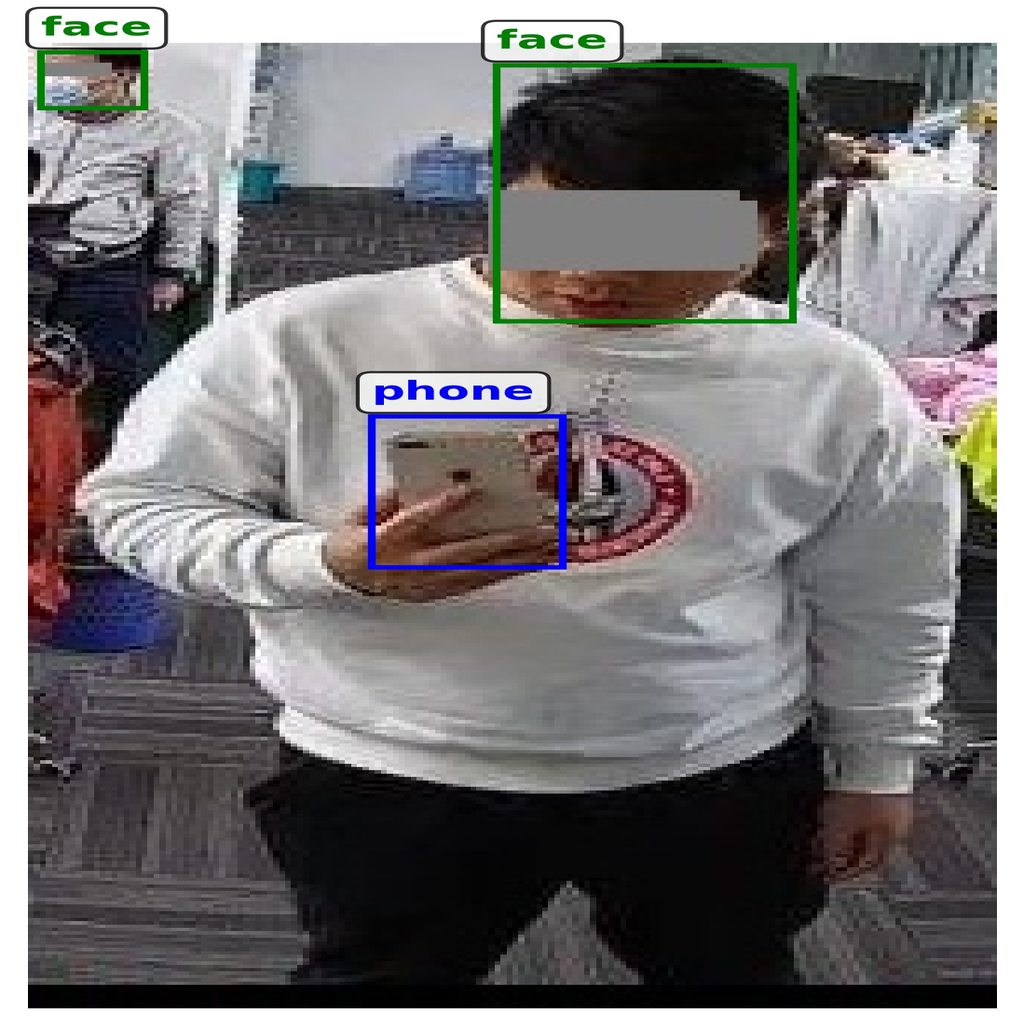}{YOLOv11-x}\hfill
\colthree{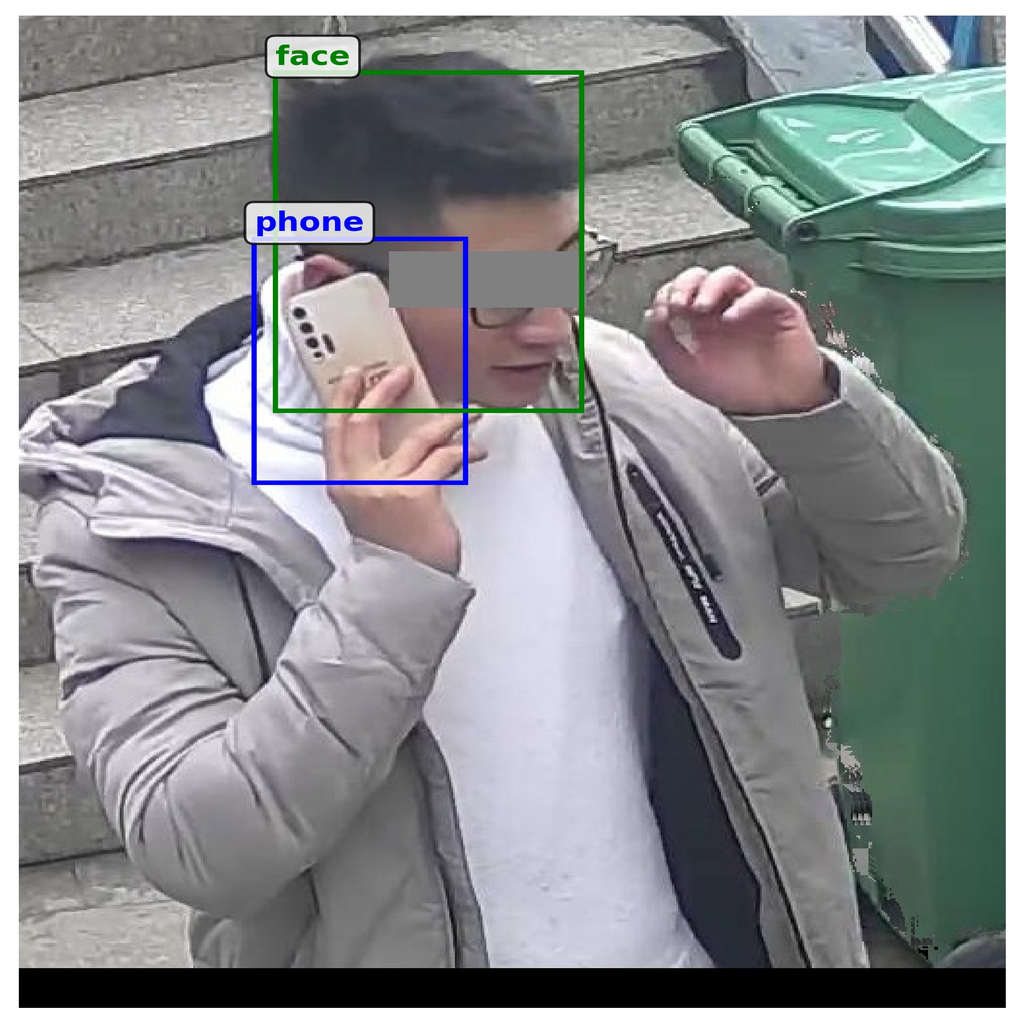}{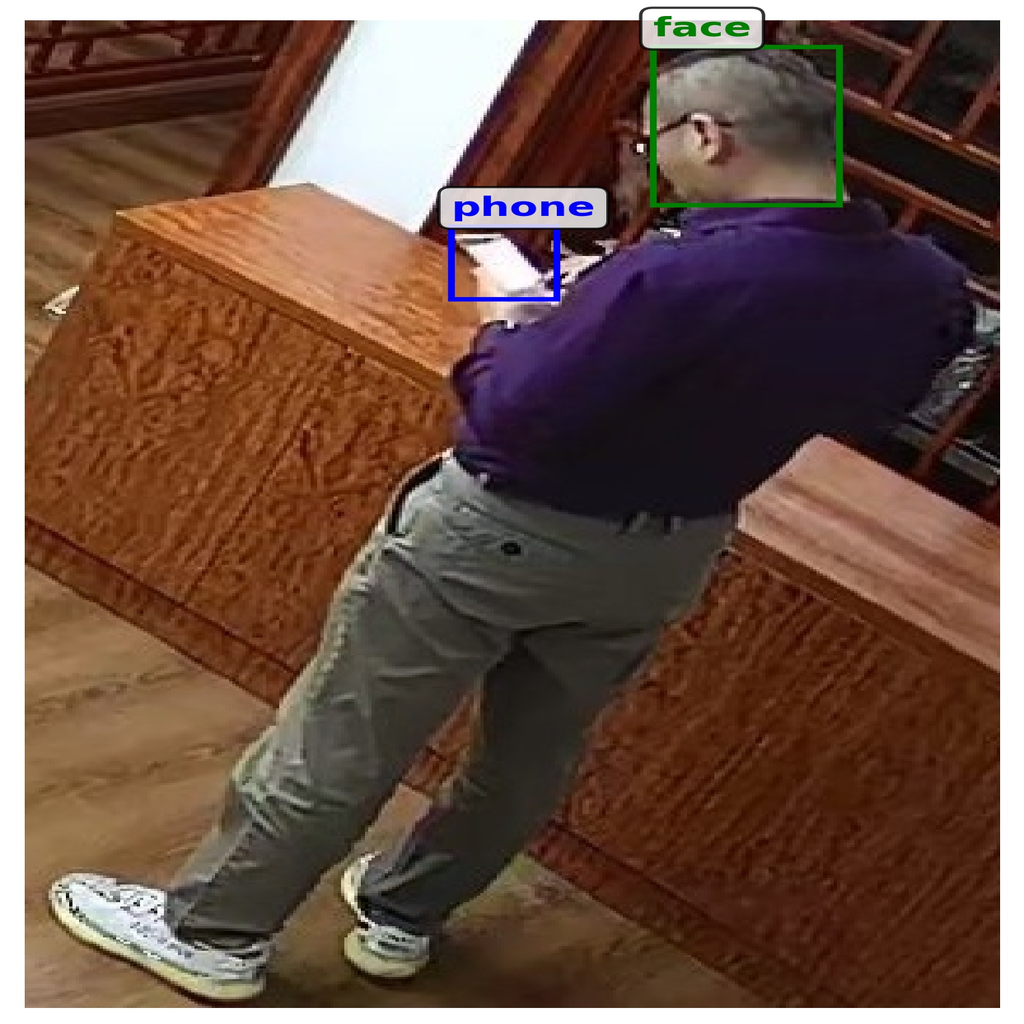}{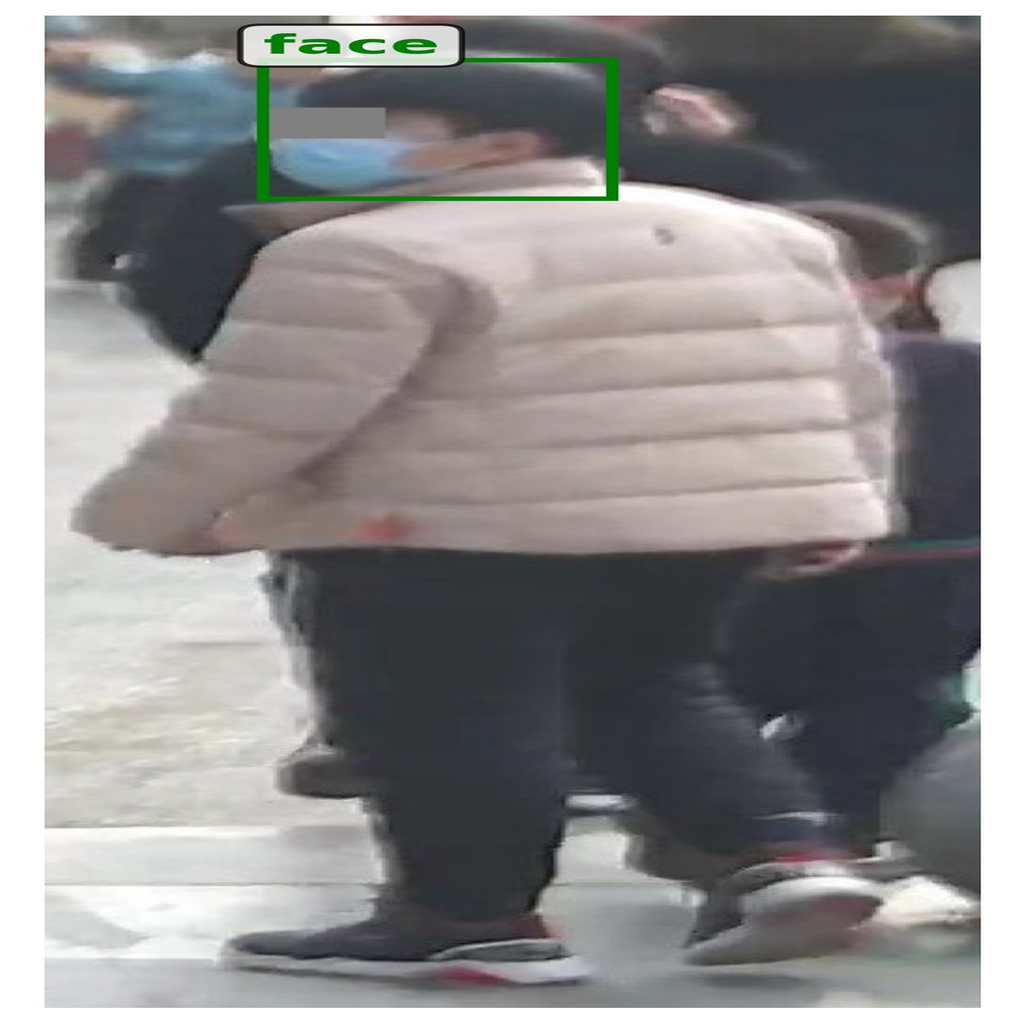}{DETR}\hfill
\colthree{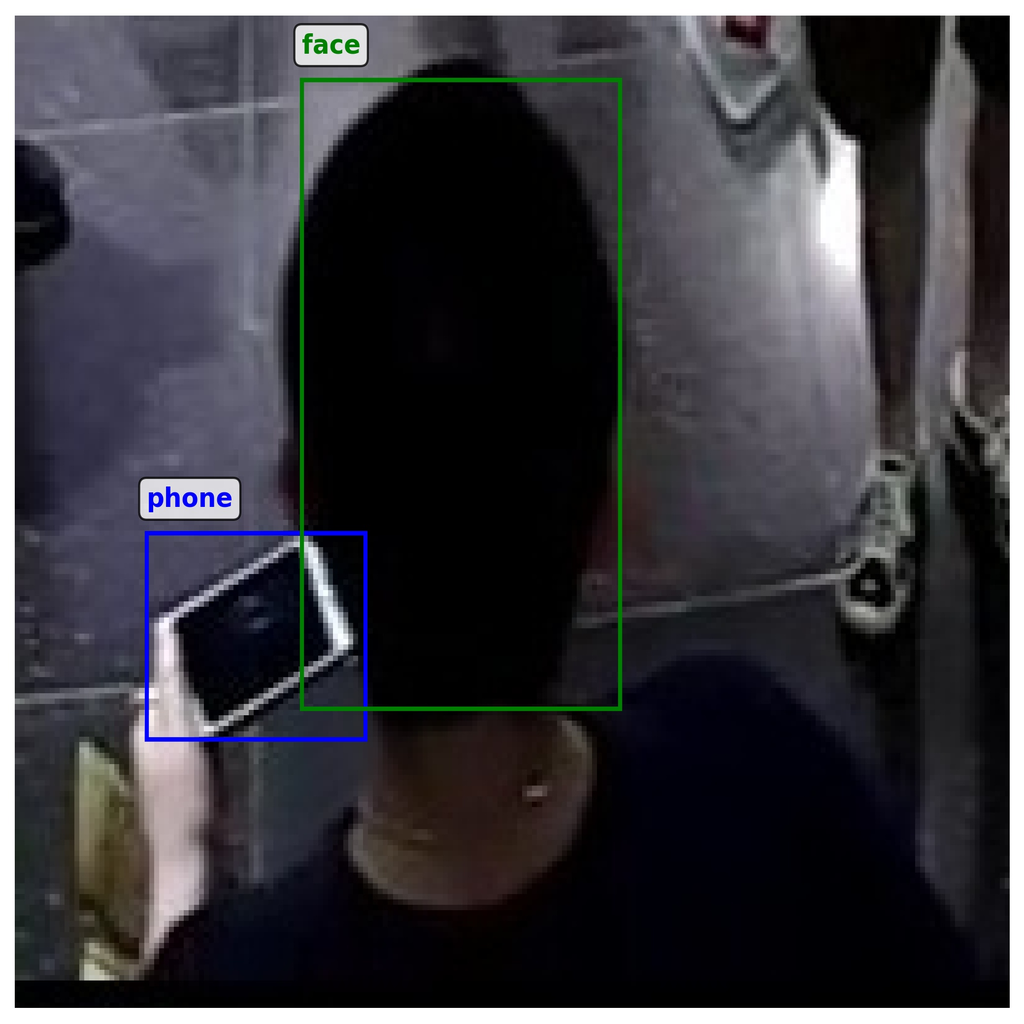}{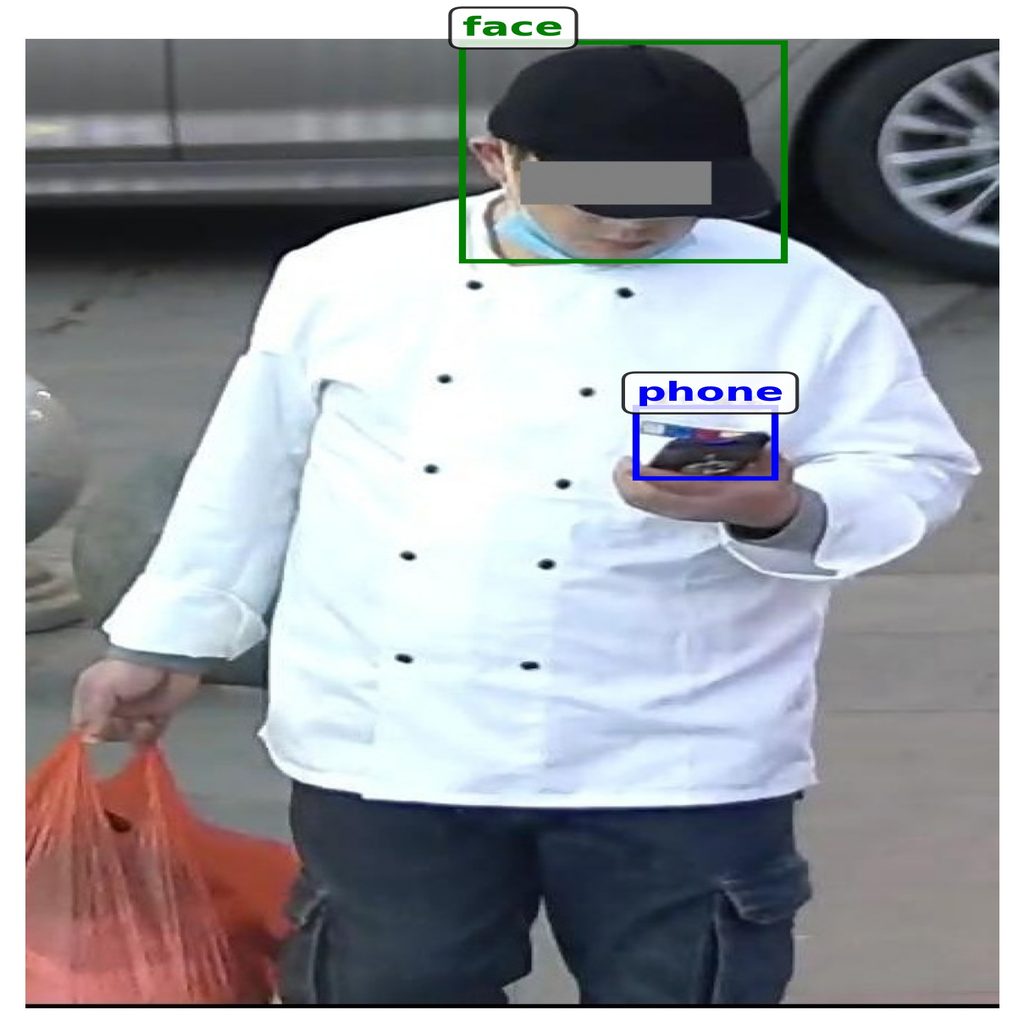}{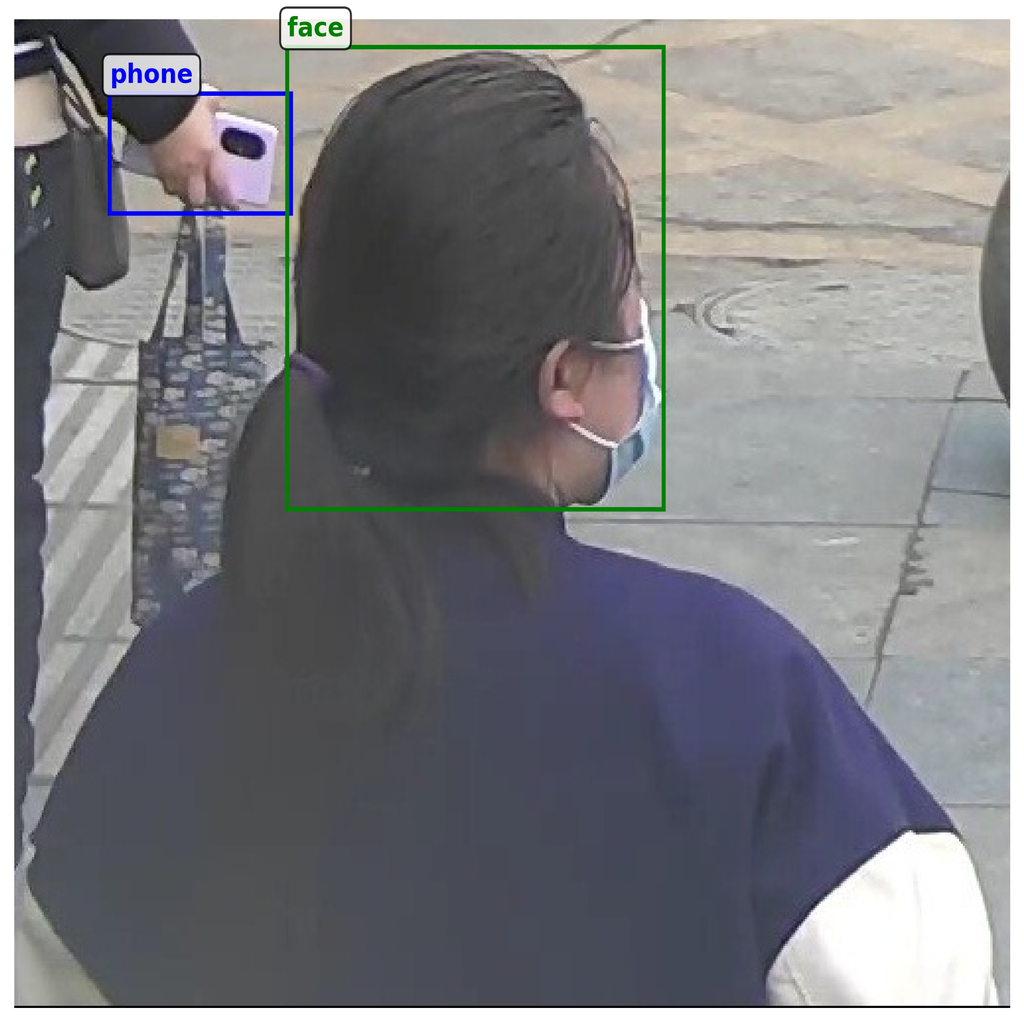}{Deform DETR}

\caption{Per-model qualitative detection examples (3×9).}
\label{fig:three_by_nine_grid}
\end{figure*}

\subsection{Results and Analysis}
The primary detection results, summarized in Table~\ref{tab:main_results}, demonstrate that all evaluated models achieve strong performance. Across YOLO variants, phone detection is consistently more accurate than face detection on our dataset. For faces, YOLOv8-x achieves the highest AP@50 at 56.5\%, while the best AP@95 is 35.8\%, tied between YOLOv11-x and YOLOv8-x. For phones, YOLOv8-x again leads with an AP@50 of 92.4\%. This gap is mainly due to task difficulty: faces are harder in FPI-Det (dense crowds, frequent occlusions, diverse poses/illumination), whereas phones are relatively easier—hence higher AP. Nevertheless, extremely small or heavily occluded phones remain difficult. In terms of throughput, YOLOv8-n is the fastest at 416.7 FPS, surpassing the next best (YOLOv11-n) by 138.9 FPS. This disparity indicates that faces are the more challenging class in FPI-Det (dense crowds, occlusions, and diverse poses/illumination), while phones are relatively easier—hence higher AP; nevertheless, extremely small or heavily occluded phones remain difficult.

\begin{table}[!h]
  \caption{Detection results on the FPI-Det test set. (Best in each column is \textbf{bold}.)}
  \label{tab:main_results}
  \centering
  \scriptsize
  \begin{adjustbox}{width=\linewidth}
  \begin{tabular}{lcccccc}
    \toprule
    \multirow{2}{*}{Model} & \multicolumn{2}{c}{Face} & \multicolumn{2}{c}{Phone} & \multirow{2}{*}{mAP@50} & \multirow{2}{*}{FPS} \\
    \cmidrule(lr){2-3} \cmidrule(lr){4-5}
    & AP@.5 & AP@.95 & AP@.5 & AP@.95 & & \\
    \midrule
    YOLOv8-n     & 54.9 & 34.6 & 91.0 & 66.7 & 72.9 & \textbf{416.7} \\
    YOLOv8-s     & 56.1 & 34.9 & 91.5 & 69.3 & 73.8 & 238.1 \\
    YOLOv8-x     & \textbf{56.5} & \textbf{35.8} & \textbf{92.4} & \textbf{71.0} & \textbf{74.5} & 54.6 \\
    YOLOv11-n    & 54.7 & 34.6 & 91.0 & 65.9 & 72.9 & 277.8 \\
    YOLOv11-s    & 54.7 & 34.8 & 89.8 & 66.0 & 72.2 & 196.1 \\
    YOLOv11-m    & 56.2 & 35.5 & 91.9 & 70.3 & 74.0 & 109.9 \\
    YOLOv11-x    & 56.1 & \textbf{35.8} & 91.7 & 69.8 & 73.9 & 52.9 \\
    \midrule
    DETR ~\cite{carion2020end}       & 39.5 & 21.2 & 57.3 & 26.9 & 48.4 & 43.0 \\
    Deform DETR~\cite{zhu2020deformable}   & 45.4 & 29.7 & 68.9 & 40.6 & 57.2 & 50.2 \\
    \bottomrule
  \end{tabular}
  \end{adjustbox}
\end{table}

Across our benchmarks, Deformable DETR consistently outperforms DETR without sacrificing speed. For the \emph{Face} class, it raises AP@0.5 from \(39.5\) to \(45.4\) and AP@.95 from \(21.2\) to \(29.7\); for \emph{Phone}, AP@0.5 improves from \(57.3\) to \(68.9\) and AP@\((0.95)\) from \(26.9\) to \(40.6\). These gains lift overall mAP@50 from \(48.4\) to \(57.2\), while throughput increases from \(43.0\) to \(50.2\) FPS.
The larger margins at the stringent IoU threshold of \(0.95\) indicate substantially tighter localization for Deformable DETR, whereas vanilla DETR remains less precise under high-overlap criteria.

\subsection{Phone Usage Behavior Classification}

Beyond detection, we add a parameter-free binary classifier that infers phone-use behavior from detector outputs, labeling an instance as “phone-in-use” when a face and a phone are co-detected in the same frame and “not-in-use” otherwise.

We formulate phone-usage detection as a binary image classification task with labels 0 (phone use) and 1 (no phone use). We evaluate with Accuracy, Precision, Recall, F1-score, and Specificity. For Precision, Recall, and F1, the positive class is “phone use” (label 0); Specificity is reported as the true negative rate for the “no phone use” class (label 1). Accuracy summarizes overall correctness. Precision quantifies how often “use” predictions are correct, controlling false alarms; Recall measures how many true “use” instances are recovered, controlling misses; F1 provides a single balance of Precision and Recall when error costs or class priors are uncertain. Specificity complements Recall by characterizing performance on the negative class, which is essential to avoid over-flagging benign scenes as phone use.

\begin{table}[!h]
\caption{Phone usage behavior classification performance metrics. (Best in each column is \textbf{bold}.)}
\label{tab:classification_results}
\centering
\scriptsize
\begin{adjustbox}{width=\linewidth}
\begin{tabular}{lccccc}
\toprule
Method & Accuracy(\%) & Precision(\%) & Recall(\%) & F1-Score(\%) & Specificity(\%)\\
\midrule
YOLOv8-n      & 88.0 & 81.8 & 90.8 & 86.0 & 86.2 \\
YOLOv8-s      & 88.4 & 82.7 & 90.1 & 86.3 & 87.2 \\
YOLOv8-x      & 89.2 & 84.9 & 89.2 & 87.0 & 89.2 \\
YOLOv11-n     & 88.9 & 85.6 & 87.2 & 86.4 & 90.0 \\
YOLOv11-s     & 87.4 & 79.8 & \textbf{92.2} & 85.5 & 84.0 \\
YOLOv11-m     & 88.9 & \textbf{86.2} & 86.3 & 86.3 & \textbf{90.6} \\
YOLOv11-x     & \textbf{89.5} & 85.9 & 89.1 & \textbf{87.5} & 89.8 \\
\midrule
DETR~\cite{carion2020end}          & 89.2 & 84.9 & 89.3 & 87.0 & 89.4 \\
Deform DETR~\cite{zhu2020deformable} & 88.4 & 83.7 & 86.9 & 85.3 & 89.3 \\
\bottomrule
\end{tabular}
\end{adjustbox}
\end{table}

The classification results presented in Table~\ref{tab:classification_results} demonstrate robust performance across all evaluation metrics. Across YOLO baselines, performance is tightly clustered, with accuracy ranging from 87.4–89.5\% and F1 from 85.5–87.5\%. YOLOv11-x offers the best overall balance—highest accuracy (89.5\%) and F1 (87.5\%), alongside strong specificity (89.8\%). Within the YOLOv8 family, YOLOv8-x is strongest (accuracy 89.2\%, F1 87.0\%) with balanced precision/recall (84.9/89.2). If sensitivity is prioritized, YOLOv11-s attains the highest recall (92.2\%) but at the cost of precision (79.8\%) and specificity (84.0\%). Conversely, YOLOv11-m is the most conservative, achieving the highest specificity (90.6\%) and top precision (86.2\%) with a balanced F1 (86.3\%). Compact variants (YOLOv8-n, YOLOv11-n) remain competitive (F1 86.0–86.4\%) with moderate specificity (86.2–90.0\%), making them attractive when efficiency is a priority. Overall, we recommend YOLOv11-x when accuracy is paramount, YOLOv11-s for recall-critical monitoring, and YOLOv11-m when minimizing false positives is more critical.

For the transformer baselines, although DETR lags behind YOLO and Deformable DETR in detection accuracy (Table~\ref{tab:main_results}), its classification metrics are on par with the best YOLO variants (accuracy 89.2\%, F1 87.0\%) and very close to Deformable DETR (88.4\%, 85.3\%). This indicates that once both a face and a phone are detected, their bounding boxes are precise enough for the co-detection rule to make a reliable binary decision; the behavior label depends mainly on correct co-occurrence rather than sub-pixel localization. The small gap between DETR and Deformable DETR in classification further suggests that finer localization helps detection AP but has limited impact on this parameter-free usage classifier, whereas detector recall chiefly governs how many frames are eligible for classification.

Qualitative analysis, illustrated in Figure~\ref{fig:three_by_nine_grid}, shows that workplace environments with stable lighting yield the highest accuracy, while transportation scenes are the most challenging. Common failure cases include false positives on tablet-like objects and false negatives on heavily occluded or very small targets, providing clear directions for future improvements.

\section{Conclusions}

We introduce FPI-Det, a new dataset of 22,879 images that, to our knowledge, is the first large-scale benchmark with joint face and phone annotations. We define the associated tasks and evaluation protocols for both object detection and phone-use behavior classification. All faces are de-identified to protect privacy. The images were collected from \url{https://www.datafountain.cn/competitions/506} and had got permissions. We benchmark mainstream detectors on FPI-Det and further propose a non‑parametric classification scheme that infers phone use directly from detector outputs.

Looking ahead, we focus on three directions: tighter attribution that links each detected phone to the correct face to reduce false positives; robust handling of borderline cases--such as holding without active use--via richer interaction cues and metric-space modeling; and field deployment of FPI-Det models on industrial cameras and at traffic intersections.

\clearpage

\bibliographystyle{IEEEbib}
\bibliography{strings,refs}

\end{document}